# Human-aided Multi-Entity Bayesian Networks Learning from Relational Data

**Cheol Young Park** CPARKF@MASONLIVE.GMU.EDU
**Kathryn Blackmond Laskey** KLASKEY@GMU.EDU
*The Sensor Fusion Lab & Center of Excellence in C4I*
*George Mason University, MS 4B5*
*Fairfax, VA 22030-4444 U.S.A.*

## Abstract

An Artificial Intelligence (AI) system is an autonomous system which emulates human's mental and physical activities such as Observe, Orient, Decide, and Act, called the OODA process. An AI system performing the OODA process requires a semantically rich representation to handle a complex real world situation and ability to reason under uncertainty about the situation. Multi-Entity Bayesian Networks (MEBNs) combines First-Order Logic with Bayesian Networks for representing and reasoning about uncertainty in complex, knowledge-rich domains. MEBN goes beyond standard Bayesian networks to enable reasoning about an unknown number of entities interacting with each other in various types of relationships, a key requirement for the OODA process of an AI system. MEBN models have heretofore been constructed manually by a domain expert. However, manual MEBN modeling is labor-intensive and insufficiently agile. To address these problems, an efficient method is needed for MEBN modeling. One of the methods is to use machine learning to learn a MEBN model in whole or in part from data. In the era of Big Data, data-rich environments, characterized by uncertainty and complexity, have become ubiquitous. The larger the data sample is, the more accurate the results of the machine learning approach can be. Therefore, machine learning has potential to improve the quality of MEBN models as well as the effectiveness for MEBN modeling. In this research, we study a MEBN learning framework to develop a MEBN model from a combination of domain expert's knowledge and data. To evaluate the MEBN learning framework, we conduct an experiment to compare the MEBN learning framework and the existing manual MEBN modeling in terms of development efficiency.

**Keywords:** Bayesian Networks, Multi-Entity Bayesian Networks, Human-aided Machine Learning

## 1 Introduction

An Artificial Intelligence (AI) system is an autonomous system which emulates human's mental and physical activities such as the OODA process [Boyd, 1976][Boyd, 1987]. The OODA process contains four steps (*Observe*, *Orient*, *Decide*, and *Act*). In the Observe step, data or signal from every mental/physical situation (e.g., states, activities, and goals) of external systems (e.g., an adversary) as well as internal systems (e.g., a command center or an allied army) in the world are observed in some internal observing guidance or control, and observations derived from data or signal are produced. In the Orient step, observations become information, formed as a model, by reasoning, analysis, and synthesis influenced from knowledge, belief, condition, etc. The Orient step can produce plan and COA (Course of Actions). Hypotheses or alternatives for models can be decided by an AI in the Decide step. In the Act step, all decided results are implemented, and real activities and states can be operated and produced, respectively. The four steps continue until the end of the life cycle of the AI system.

An AI system performing the OODA process requires a semantically rich representation to handle situations in a complex real and/or cyber world. Furthermore, the number of entities and the relationships among them may be uncertain. For this reason, the AI system needs an expressive formal language for representing and reasoning about uncertain, complex, and dynamic situations. Multi-Entity Bayesian Networks (MEBNs) [Laskey, 2008] combines First-Order Logic with Bayesian Networks (BNs) [Pearl, 1988] for representing and reasoning about uncertainty in complex, knowledge-rich domains. MEBN goes beyond standard Bayesian networks to enable reasoning about an unknown number of entities interacting with each other in various types of relationships, a key requirement for the AI system.

MEBN has been applied to AI systems [Laskey et al., 2000][Wright et al., 2002][Costa et al., 2005][Suzic, 2005][Costa et al., 2012][Park et al., 2014][Golestan, 2016][Li et al., 2016][Park et al., 2017]. In a recent review of knowledge representation formalisms for AI, Golestan et al. [2016] recommended MEBN as having the most comprehensive coverage of features needed to represent complex situations. Patnaikuni et al., [2017] reviewed various applications using MEBN.

In previous applications of MEBN to the AI system, the MEBN model or MTheory was constructed manually by a domain expert using a MEBN modeling process such as Uncertainty Modeling Process for Semantic Technology (UMP-ST) [Carvalho et al., 2016]. Manual MEBN modeling is a labor-intensive and insufficiently agile process. Greater automation through machine learning may save labor and enhance agility. For this reason, Park et al. [2016] introduced a process model called Human-aided Multi-Entity Bayesian Networks Learning for Predictive Situation Awareness by combining domain expertise with data. The process model was focused on the predictive situation awareness (PSAW) domain. However, the process model is not necessary to be only applied to the PSAW domain. This paper defines a general process model for *Human-aided Multi-Entity Bayesian Networks Learning*[1] called *HML*. HML specifies four steps with guidelines about how to associate with (1) domain knowledge, (2) database model, and (3) MEBN learning. Thus, the general process model is capable of generalization to reuse a variety of domains to develop a domain specific MEBN model (e.g., predictive situation awareness, planning, natural language processing, and system modeling). (1) Domain knowledge can be specified by a reference model which is an abstract framework to which a developer refers in order to develop a specific model. Such a reference model can support the design of a MEBN model in the certain domain and improve the quality of the MEBN model. (2) A database model can support to the design of a MEBN model for automation, if there are common elements between the database model and MEBN model. For example, Relational Model (RM), which is a database model based on first-order predicate logic [Codd, 1969; Codd, 1970] and the most widely used data model in the world, represent entities and attributes. Such entities and attributes in RM can be mapped to entities and random variables in MEBN, respectively. Thus, common elements between a database model and MEBN can be used to automated conversion. In this research, we introduce the use of RM as the database model for MEBN learning. (3) MEBN learning is to learn an optimal MEBN model which fits well an observed datasets in database models. MEBN learning can be classified into two types: One is MEBN structure learning (e.g., finding optimal structures of MEBN) and another is MEBN parameter learning (e.g., finding an optimal set of parameters for local distributions of random variables in MEBN). In this research, MEBN parameter learning is introduced. Overall, HML contains three supportive methodologies: (1) a domain reference model (e.g., a reference model for predictive situation awareness, planning, natural language processing, or system modeling), (2) a mapping between MEBN and a database model (e.g., RM), and (3) MEBN learning (e.g., a conditional linear Gaussian parameter learning for MEBN) to develop a MEBN model efficiently and effectively. In this research, we conduct an experiment to compare Human-aided Multi-Entity Bayesian Networks Learning (HML) and the existing manual MEBN modeling in terms of development efficiency.

---

[1] This paper is an extension of the conference paper, [Park et al., 2016].

Section 2 provides background information about MEBN and an existing MEBN modeling process. In Section 3, a relational database, which is an illustrative database model used to explain HML, is introduced. In Section 4, HML is presented with an illustrative example. In Section 5, an experiment comparing between HML and the existing MEBN modeling process is introduced. Finally, conclusions are presented and future research directions are discussed.

## 2 Background

This section provides background information about Multi-Entity Bayesian Networks (MEBNs), a script form of MEBN, and Uncertainty Modeling Process for Semantic Technology (UMP-ST). In Section 2.1, MEBN as a representation formalism is presented with some definitions and an example for MEBN. In Section 2.2, a simple script form of MEBN is introduced. HML in this research is a modification of UMP-ST, so UMP-ST is introduced in Section 2.3.

### 2.1 Multi-Entity Bayesian Network

In this section, we describe MEBN and a graphical representation for MEBN. Details can be found in [Laskey, 2008]. The following definitions are taken from [Laskey, 2008]. MEBN allows compact representation of repeated structure in a joint distribution on a set of random variables. In MEBN, random variables are defined as templates than can be repeatedly instantiated to construct probabilistic models with repeated structure. MEBN represents domain knowledge using an MTheory, which consists of a collection of MFrags (see Fig. 1). An MFrag is a fragment of a graphical model that is a template for probabilistic relationships among instances of its random variables. Random variables (RVs) may contain ordinary variables, which can be instantiated for different domain entities. We can think of an MFrag as a class which can generate instances of BN fragments. These can then be assembled into a Bayesian network, called a *situation-specific Bayesian Network* (SSBN), using an SSBN algorithm [Laskey, 2008]. In other words, a given MTheory can be used to construct many different SSBNs for different situations.

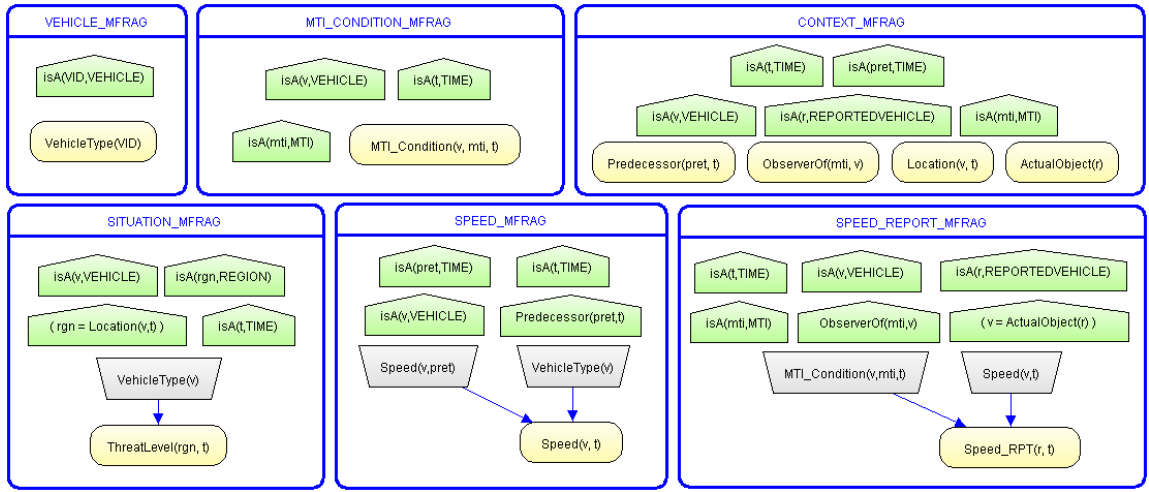

**Fig. 1 Threat Assessment MTheory or MEBN Model**

To understand how this works, consider Fig. 1, which shows an MTheory called the *Threat*

*Assessment* MTheory. This MTheory contains six MFrags: *Vehicle*, *MTI_Condition*, *Context*, *Situation*, *Speed*, and *Speed_Report*. An MFrag (e.g., Fig. 2) may contain three types of random variables: *context* RVs, denoted by green pentagons, *resident* RVs, denoted by yellow ovals, and *input* RVs, denoted by gray trapezoids. Each MFrag defines local probability distributions for its input RVs. These distributions may depend on the input RVs, whose distributions are defined in other MFrags. Context RVs express conditions that must be satisfied for the distributions defined in the MFrag to apply.

Specifically, Fig. 2 shows the *Situation* MFrag (from the *Threat Assessment* MTheory) used for an illustrative example of an MFrag. The *Situation* MFrag represents probabilistic knowledge of how the threat level of a region at a time is measured depending on the vehicle type of detected objects. For example, if in a region there are many tracked vehicles (e.g., Tanks), the threat level of the region will be high. An MFrag consists of a set of resident nodes, a set of context nodes, a set of input nodes, an acyclic directed graph for the nodes, and a set of class local distributions (CLD) for the nodes. The context nodes (i.e., *isA*(*v*, *VEHICLE*), *isA*(*rgn*, *REGION*), *isA*(*t*, *TIME*), and *rgn = Location*(*v*, *t*)) for this MFrag (shown as pentagons in the figure) show that this MFrag applies when a vehicle entity is substituted for the ordinary variable *v*, a region entity is substituted for the ordinary variable *rgn*, a time entity is substituted for the ordinary variable *t*, and a vehicle *v* is located in region *rgn* at time *t*. The context node *rgn = Location*(*v*, *t*) constrains the values of *v*, *rgn*, and *t* from the possible instances of vehicle, region, and time, respectively. For example, suppose *v1* and *v2* are vehicles and *r1* is a region in which the only *v1* is located at time *t1*. The context node *rgn = Location*(*v*, *t*) will allow only an instance of (*v1*, *r1*, *t1*) to be selected, but not (*v2*, *r1*, *t1*), because *r1* is not the location of *v2* at *t1*. Next, we see the input node *VehicleType*(*v*), depicted as a trapezoid. Input nodes are nodes whose distribution is defined in another MFrag. For example, a resident node *VehicleType*(*v*) is found in the MFrag *Vehicle* from the top left in Fig. 1.

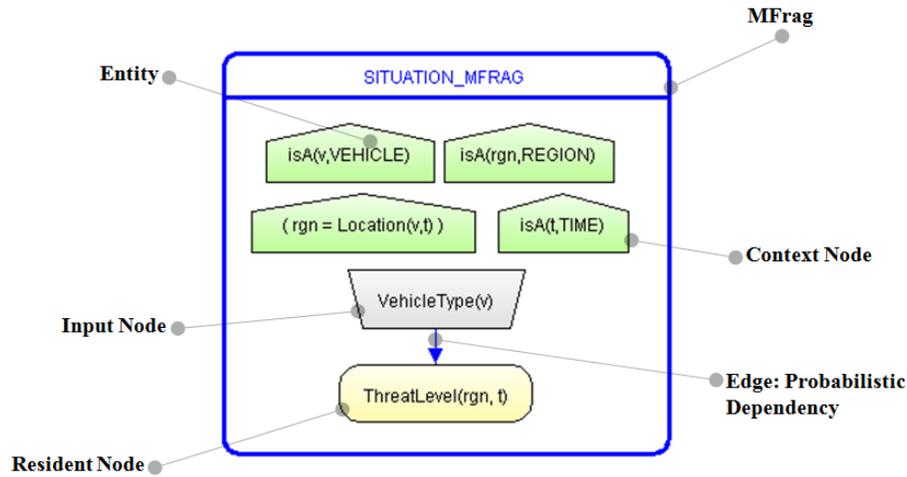

Fig. 2 Situation MFrag

In Fig. 2, the node *ThreatLevel*(*rgn*, *t*) is a resident node, which means its distribution is defined in the MFrag of the figure. Like the graph of a common BN, the fragment graph shows probabilistic dependencies. CLD 2.1 in the script below shows that a class local distribution for *ThreatLevel*(*rgn*, *t*) describes its probability distribution as a function of the input nodes given the instances that satisfy the context nodes. The *class local distribution* (CLD) $L^C$ can be used to produce an *instance local distribution* (ILD) $L^I$ in a SSBN. Note that in Subsection 4.3.2, these CLD and ILD are defined formally. In this Subsection, we introduce CLD with a simple illustrative example. The class local distribution of *ThreatLevel*(*rgn*, *t*), which depends on the type of vehicle, can be expressed as CLD 2.1. The CLD is defined in a language called Local Probability Description Language (LPDL). In our example, the probabilities of the states, High and Low, of *ThreatLevel*(*rgn*, *t*) are defined as a function of the values, High and Low, of instances *rgn* = *Location*(*v*, *t*) of the parent nodes that satisfy the context constraints. For the high state in the first if-scope in CLD 2.1, the probability value is assigned by the function described by "1 – 1 / CARDINALITY(*v*)". The CARDINALITY function returns the number of instances of *v* satisfying the if-condition. For example, in CLD 2.1, if the situation involves three vehicles and two of them are tracked, then the CARDINALITY function will return 2. We see that as the number of tracked vehicles becomes very large, the function, "1 – 1 / CARDINALITY(*v*)", will tend to 1. This means the threat level of the region will be very high.

**CLD 2.1**: The class local distribution for the resident node *ThreatLevel*(*rgn*, *t*)

```
1   if some v have (VehicleType = Tracked) [
2       High = 1 – 1 / CARDINALITY(v),
3       Low = 1 – High
4   ] else [
5       High = 0,
6       Low = 1
7   ]
```

Alternatively, we might model the resident node *ThreatLevel*(*rgn*, *t*) as a continuous random variable. For a continuous resident node, the class local distribution is defined by a continuous probability density function. The class local distribution of the continuous resident node (see CLD 2.2) can be described by LPDL, also.

**CLD 2.2**: The class local distribution for the continuous resident node *ThreatLevel*(*rgn*, *t*)

```
1   if some v have (VehicleType = Tracked) [
2       10 * CARDINALITY(v) + NormalDist(10, 5)
3   ] else [
4       NormalDist(10, 5)
5   ]
```

The meaning of CLD 2.2 is that the degree of the threat in the region is 10 * the number of tracked vehicles plus a normally distributed error with mean 10 and variance 5. Currently, LPDL limits continuous nodes to conditional linear Gaussian (CLG) distributions [Sun et al., 2010], defined as:

$$\text{p}(R \mid \text{Pa}(R), CF_j) = \mathcal{N}(m + b_1 P_1 + b_2 P_2 \ldots, +b_n P_n, \sigma^2), \qquad (2.1)$$

where Pa() is a set of continuous parent resident nodes of the continuous resident node, $R$, having $\{P_1, \ldots, P_n\}$, $P_i$ is a $i$-th continuous parent node, $CF_j$ is a $j$-th configuration of the discrete parents of $R$ (e.g., $\mathbf{CF} = \{CF_1 = (VehicleType = Tracked), CF_2 = (VehicleType = Wheeled)\}$), $m$ is a regression intercept, $\sigma^2$ is a conditional variance, and $b_i$ is regression coefficient.

Using the above MTheory example, we define elements of MTheory more precisely. The following definitions are taken from [Laskey, 2008].

**Definition 2.1 (MFrag)** An MFrag *F*, or MEBN fragment, consists of: (*i*) a set ***C*** of context nodes, which represent conditions under which the distribution defined in the MFrag is valid; (*ii*) a set ***I*** of input nodes, which have their distributions defined elsewhere and condition the distributions defined in the MFrag; (*iii*) a set ***R*** of resident nodes, whose distributions are defined in the MFrag[2]; (*iv*) an acyclic directed graph ***G***, whose nodes are associated with resident and input nodes; and (*iv*) a set $\boldsymbol{L}^C$ of class local distributions, in which an element of $\boldsymbol{L}^C$ is associated with each resident node.

The nodes in an MFrag are different from the nodes in a common BN. A node in a common BN represents a single random variable, whereas a node in an MFrag represents a collection of RVs: those formed by replacing the ordinary variables with identifiers of entity instances that meet the context conditions. To emphasize the distinction, we call the resident nodes in the MBEN nodes, or MNodes.

MNodes correspond to predicates (for true/false RVs) or terms (for other RVs) of first-order logic. An MNode is written as a predicate or term followed by a parenthesized list of ordinary variables as arguments.

**Definition 2.2 (MNode)** An *MNode*, or MEBN Node, is a random variable *N*(*ff*) specified an *n*-ary function or predicate of first-order logic (FOL), a list of *n* arguments consisting of ordinary variables, a set of mutually exclusive and collectively exhaustive possible values, and an associated class local distribution. The special values *true* and *false* are the possible values for predicates, but may not be possible values for functions. The RVs associated with the MNode are constructed by substituting domain entities for the *n* arguments of the function or predicate. The class local distribution specifies how to define local distributions for these RVs.

For example, the node *ThreatLevel*(*rgn*, *t*) in Fig. 2 is an MNode specified by a FOL function *ThreatLevel*(*rgn*, *t*) having two possible values (i.e., *High* and *Low*).

**Definition 2.3 (MTheory)** An *MTheory M*, or MEBN Theory, is a collection of MFrags that satisfies conditions given in [Laskey, 2008] ensuring the existence of a unique joint distribution over its random variables.

An MTheory is a collection of MFrags that defines a consistent joint distribution over random variables describing a domain. The MFrags forming an MTheory should be mutually consistent. To ensure consistency, conditions must be satisfied such as no-cycle, bounded causal depth, unique home MFrags, and recursive specification condition [Laskey, 2008]. No-cycle means that the generated SSBN will contain no directed cycles. Bounded causal depth means that depth from a root node to a leaf node of an instance SSBN should be finite. Unique home MFrags means that each random variable has its distribution defined in a single MFrag, called its home MFrag. Recursive specification means that MEBN provides a means for defining the distribution for an RV depending on an ordered ordinary variable from previous instances of the RV.

The *IsA* random variable is a special RV representing the type of an entity. IsA is commonly used as a context node to specify the type of entity that can be substituted for an ordinary variable in an MNode.

**Definition 2.4 (IsA random variable)** An *IsA random variable*, IsA(*ov*, *tp*), is an RV corresponding to a 2-argument FOL predicate. The IsA RV has value *true* when its second argument *tp* is filled by the type of its first argument *ov* and *false* otherwise.

For example, in the *Situation* MFrag in Fig. 2, *isA*(*v*, *VEHICLE*) is an *IsA* RV. Its first argument *v* is filled by an entity instance and its second argument is the type symbol *VEHICLE*. It has value

---

[2] Bold italic letters are used to denote sets.

*true* when its first argument is filled by an object of type *VEHICLE*.

## 2.2 Script for MEBN

Fig. 1 shows a graphical representation for an MTheory. In this subsection, we introduce a script representing an MTheory. This script is useful to manage contents of an MTheory. The *Threat Assessment* MTheory in Fig. 1 can be represented by the following script (MTheory 2.1).

---

**MTheory 2.1**: Part of Script MTheory for Threat Assessment

```
1   [F: Situation
2       [C: IsA (v, VEHICLE)][C: IsA (rgn, REGION)][C: IsA (t, TIME)]
3       [C: rgn = Location (v, t)]
4       [R: ThreatLevel (r, t)
5           [IP: VehicleType (v)]
6       ]
7   ]
8   [F: Vehicle
9       [C: IsA (vid, VEHICLE)]
10      [R: VehicleType (vid)]
11  ]
12  …
```

---

The script contains several predefined single letters (F, C, R, IP, RP, and L). The single letters, F, C, and R denote an MFrag, a context node, and a resident node, respectively. For a resident node (e.g., *Y*) in an MFrag, a resident parent (RP) node (e.g., *X*), which is defined in the MFrag, is denoted as RP (e.g., [R: *Y* [RP: *X*]]). For an input node, we use a single letter IP. Each node can contain a CLD denoted as L. For example, suppose that there is a CLD type called *ThreatLevelCLD*. If the resident node *ThreatLevel* in Line 4 uses the CLD type *ThreatLevelCLD*, the resident node *ThreatLevel* can be represented as [R: ThreatLevel (*rgn*, *t*) [L: *ThreatLevelCLD*]].

## 2.3 Uncertainty Modeling Process for Semantic Technology (UMP-ST)

Traditional ontologies [Smith, 2003] are limited to deterministic knowledge. Probabilistic Ontologies (POs) move beyond this limitation by incorporating formal probabilistic semantics. Probabilistic OWL (PR-OWL) [Costa, 2005] is a probabilistic ontology language that extends OWL with semantics based on Multi-Entity Bayesian Networks (MEBNs), a Bayesian probabilistic logic [Laskey, 2008]. PR-OWL has been extended to PR-OWL 2 [Carvalho et al., 2017], which provides a tighter link between the deterministic and probabilistic aspects of the Ontologies. Developing probabilistic ontologies can be greatly facilitated by the use of a modeling framework such as the UMP-ST [Carvalho et al., 2016]. UMP-ST was applied for construction of PR-OWL 1 & 2 probabilistic ontologies. The UMP-ST process consists of four main disciplines: (1) *Requirement*, (2) *Analysis & Design*, (3) *Implementation*, and (4) *Test*.

(1) The *Requirement* discipline defines goals, queries, and evidence for a probabilistic ontology. The goals are objectives to be achieved by reasoning with the probabilistic ontology (e.g., identify a ground target). The queries are specific questions for which the answers help to achieve the objectives. For example, what is the type of the target? The evidence is information used to answer the query (e.g., history of the speed of the target). (2) The *Analysis & Design* discipline designs entities, attributes for the entities, relationships between the entities, and rules for attributes and relationships to represent uncertainty. These are associated with the goals, queries, and evidence in the *Requirement* discipline. For example, suppose that a vehicle entity has two attributes, type, and speed. Then an example of a rule might be that if the speed is low, the type is likely to be a tracked vehicle. (3) The *Implementation* discipline is a step to develop a probabilistic ontology from outputs developed in the *Analysis & Design* discipline. Entities, attributes, relationships, and rules are mapped to elements of the probabilistic ontology. For example, the attributes type and speed are mapped to random variables type and speed, respectively. The rule for the speed and type is converted to the joint probability for the random variables type and speed. (4) In the *Test* discipline, the probabilistic ontology developed in the previous step is evaluated to assess its correctness. The correctness can be measured by three approaches: (a) *Elicitation Review*, in which completeness of the probabilistic ontology addressing requirements are reviewed, (b) *Importance Analysis*, which is a form of sensitivity analysis that examines the strength of influence of each random variable on other random variables, and (c) *Case-based Evaluation*, in which various scenarios are defined and used to examine the reasoning implications of the probabilistic ontology [Laskey & Mahoney, 2000].

## 3   Illustrative Running Example of Relational Data for MEBN Learning

In this section, we introduce an illustrative running example of relational data. For an illustrative example for MEBN learning, a threat assessment relational database (RDB) is introduced.

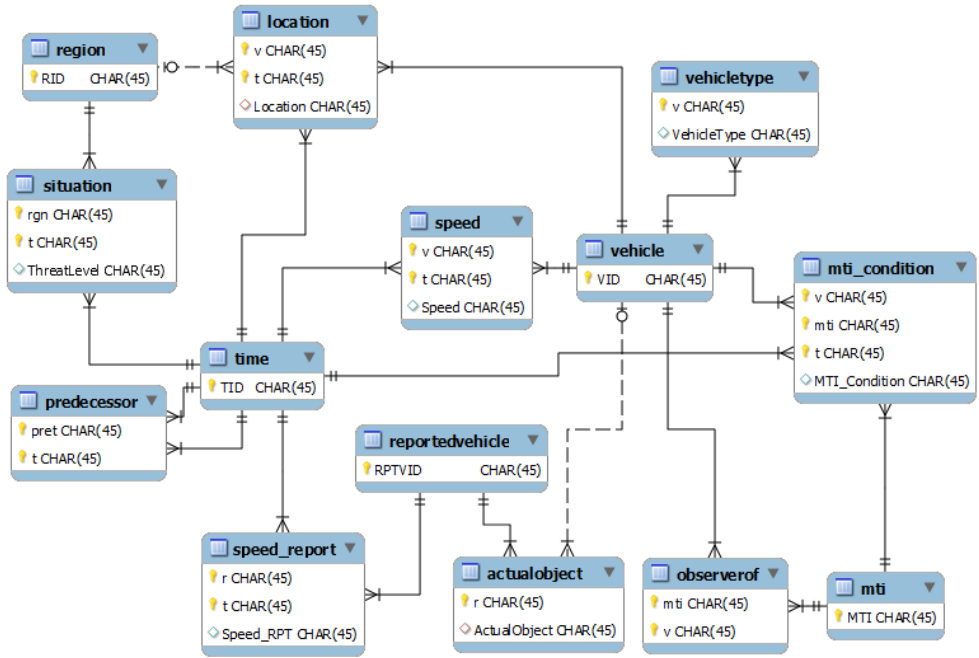

Fig. 3 Schema of a threat assessment relational database

Fig. 3 shows a schema for the threat assessment RDB. In the example RDB schema, there are 14 relations: *Region*, *Situation*, *Location*, *Time*, *Speed*, *Speed_Report*, *ActualObject*, *ObserverOf*, *Vehicle*, *VehicleType*, *Predecessor*, *ReportedVehicle*, *MTI*, and *MTI_Condition*. The relation *Region* is for region information in this situation which can contain a region index (e.g., *region1* and *region2*). The relation *Time* is for time information which is a time stamp representing a time interval (e.g., *t1* and *t2*). The relation *Vehicle* is for vehicle information which is an index of a ground-vehicle (e.g., *v1* and *v2*). The relation *VehicleType* indicates a type of the vehicle (e.g., *Wheeled* and *Tracked*). The relation *MTI* is for a moving target indicator (e.g., *mti1* and *mti2*). An MTI can be in a condition (e.g., *Good* and *Bad*) depending on weather and/or maintenance conditions. The relation *MTI_Condition* indicates the condition of an MTI. The relation *Location* is for a location where a vehicle is located. The relation *Situation* indicates a threat level to a region at a time (e.g., *Low* and *High*). The relation *ReportedVehicle* indicates a reported vehicle from an MTI. The relation *Speed* indicates an actual speed of a vehicle, while the relation *Speed_Report* indicates a reported speed of the vehicle from an MTI. The relation *ActualObject* maps a reported vehicle to an actual vehicle. The relation *ObserverOf* indicates that an MTI observes a vehicle. The relation *Predecessor* indicates a temporal order between two-time stamps.

Table 1 shows parts of the relations of the threat assessment RDB for the schema in Fig. 3. As shown Table 1, we choose six relations (*Vehicle*, *Time*, *Region*, *VehicleType*, *Location*, and *Situation*), which are used for an illustrative example through the next section. For example, the relation *Vehicle* contains a primary key **VID**. The relation *VehicleType* contains a primary key **v/Vehicle.VID**, which is a foreign key from the primary key **VID** in the relation *Vehicle* and an attribute *VehicleType*.

**Table.1 Parts of the threat assessment relational database**

| Vehicle | Time | Region | VehicleType | | Location | | | Situation | | |
|---|---|---|---|---|---|---|---|---|---|---|
| **VID** | **TID** | **RID** | **v/Vehicle.VID** | Vehicle Type | **v/Vehicle.VID** | **t/Time** | Location /Region | **Rgn /Region.RID** | **t/Time.TID** | Threat Level |
| v1 | t1 | rgn1 | v1 | Wheeled | v1 | t1 | rgn1 | rgn1 | t1 | High |
| v2 | t2 | rgn2 | v2 | Tracked | v1 | t2 | rgn1 | rgn2 | t3 | Low |
| ... | ... | ... | ... | ... | ... | ... | ... | ... | ... | ... |

## 4 Process for Human-Aided MEBN Learning

The process this research presents uses expert knowledge to define the set of possible parameters and structures. The process, called Human-aided MEBN learning (HML), modifies UMP-ST [Carvalho et al., 2016] to incorporate learning from data. As with UMP-ST, HML includes four steps (Fig. 4): (1) *Analyze Requirements*, (2) *Define World Model*, (3) *Construct Reasoning Model*, and (4) *Test Reasoning Model*.

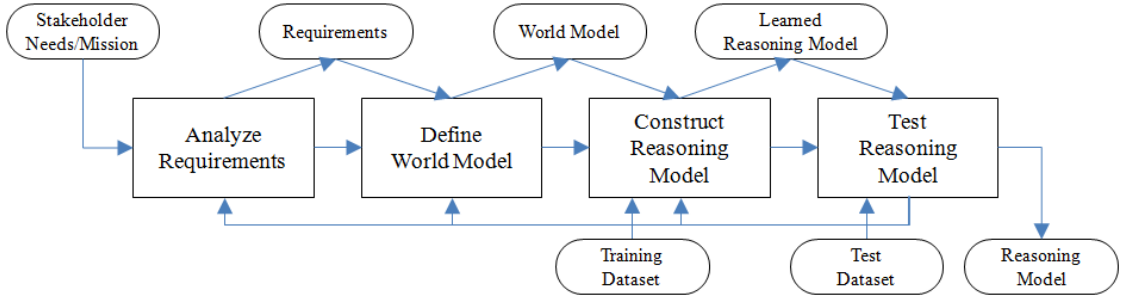

**Fig. 4 Process for Human-Aided MEBN Learning**

Initial inputs of the process can be needs and/or missions from stakeholders in a certain domain (e.g., predictive situation awareness, planning, natural language processing, and system modeling). In the *Analyze Requirements* step, specific requirements for a reasoning model (in our case, an MTheory) were identified. According to different domain type, the goals and the reasoning model will be different. For example, the goal of the reasoning model for the PSAW domain can be to identify a threatening target. The reasoning model in such domain may contain sensor models representing sensing noise. The goal of the reasoning model for the natural language processing domain can be to analyze natural languages or classify documents. The reasoning model in such domain may contain random variables specifying text corpus. In the *Define World Model* step, a target world where the reasoning model operates is defined. In the *Construct Reasoning Model* step, a training dataset can be an input for MEBN learning to learn a reasoning model. In the *Test Reasoning* step, a test dataset can be an input for the evaluation of the learned reasoning model. An output of the process is the evaluated reasoning model. The following subsections describe these four steps with the illustrative example (Section 3) of threat assessment in the PSAW domain.

## 4.1 Analyze Requirements

This step is to identify requirements for development of a reasoning model. As with requirements in UMP-ST (Section 2.3), requirements in the HML define goals to be achieved, queries to answer, and evidence to be used in answering queries. Also, the requirements should include performance criteria for verification of the reasoning model. These performance criteria are used in the *Test Reasoning Model* step. Before the *Analyze Requirements* step begins, stakeholders provide their initial requirements containing needs, wants, missions, and objectives. These initial requirements may not be defined formally. Therefore, to clarify the initial requirements, operational scenarios are developed. In other words, the operational scenarios are used to identify the goals, queries, and evidence in the requirements.

This step contains three sub-steps (Fig. 5): (1) an *Identify Goals* step, (2) an *Identify Queries/Evidence* step, and (3) a *Define Performance Criteria* step.

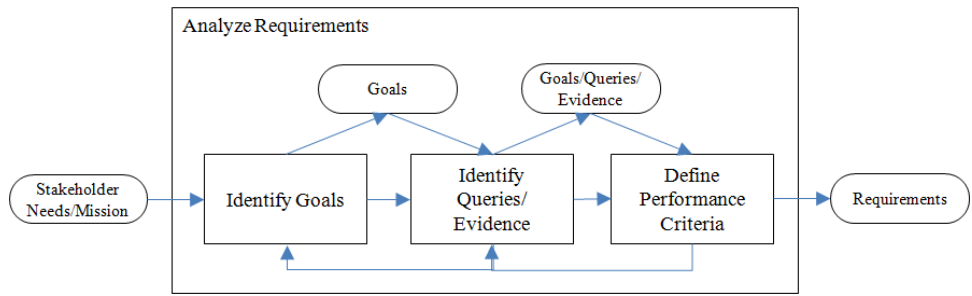

**Fig. 5 Analyze Requirements**

### 4.1.1 Identify Goals

The goals represent missions of the reasoning model we are developing. In this step, we can use a set of common questions in a certain domain, which enables us to grasp some ideas for what questions the reasoning model should answer. Such domain questions can be determined by knowledge from experts of the domain. For example, in the predictive situation awareness (PSAW) domain, several PSAW questions derived from PSAW domain knowledge can be used in this step (e.g., examples of PSAW questions can be "does a (grouped) target exist?", "what are the environmental conditions?", and "what group does the target belong to?"). Requirement 4.1 illustrates a goal we will use.

**Requirement 4.1**

**Goal 1**: Identify characteristics of a target.

### 4.1.2 Identify Queries/Evidence

The queries are specific questions for which the reasoning model is used to estimate and/or predict answers. The evidence consists of inputs used for reasoning. From these sub-steps, a set of goals, a set of queries for each goal, and a set of evidence for each query are defined. The following shows an illustrative example of defining a requirement**.**

> **Requirement 4.1**
>
> **Goal 1**: Identify characteristics of a target.
>
> > **Query 1.1**: What is the speed of the target at a given time?
> >
> > > **Evidence 1.1.1**: A speed report from a sensor.
> > >
> > > ...

### 4.1.3 Define Performance Criteria

It is necessary to evaluate whether the results for a reasoning model which will be learned from data address performance requirements in terms of reasoning. Criteria for evaluating reasoning performance include: *Speed* (e.g., execution or computation time for reasoning), *Accuracy* (e.g., measuring gap between an actual value and estimation) and *Resource Usage* (e.g., memory or CPU usage). In some situations, execution time for a reasoning model is the most important factor. In other cases, accuracy for a reasoning model may be more important. For example, an initial missile tracking may require high-speed reasoning to estimate the missile trajectory, while matching faces in a security video against a no-fly database may prioritize accuracy over execution time.

The performance criteria in the requirements can be specified in terms of some measure of accuracy (e.g., the Brier score [Brier, 1950] or the continuous ranked probability score (CRPS) [Gneiting & Raftery, 2007]). For example, we might require that the average of CRPS values between ground truth and estimated results from a reasoning model shall be less than a given threshold.

The performance criteria are determined by stakeholder agreement. Such performance criteria can be acquired through the following approaches: (1) survey, (2) experience, and (3) standard metrics drawn from published literature and standards. (1) Performance criteria can be derived by agreement of stakeholders using survey. (2) Subject matter experts can provide appropriate performance criteria from their experience. (3) Standards or literature can be used to obtain such performance criteria.

### 4.2 Define World Model

The *Define World Model* step develops a world model consisting of a structure model and rules. The world model describes a target situation of concern that is the subject. The structure model can contain entities (e.g., *target* and *sensor*), variables (e.g., *Speed* and *ThreatLevel*), and relations (e.g., *location* and *situation*). The rules describe the causal relationships between entities in the structure model (e.g., the type of a target can influence the speed of the target). The causal relationships can contain more specific information such as types of distributions and parameters for the distributions which will be used to develop an initial MTheory in the next step. The structure model and the rules provide a clear idea by which the reasoning model can be formed.

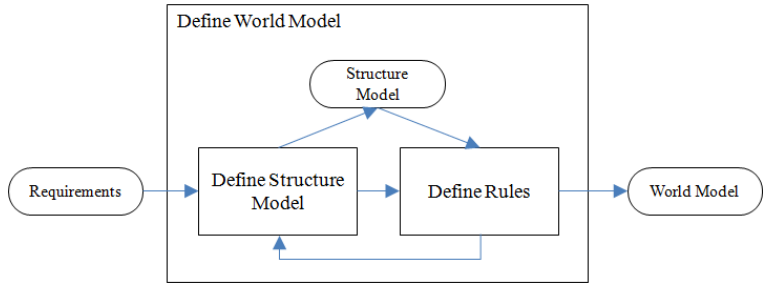

**Fig. 6 Define World Model**

This step decomposes into two sub-steps (Fig. 6): (1) a *Define Structure Model* step and (2) a *Define Rules* step. The *Define Structure Model* step defines the structure model from the requirements, domain knowledge and/or existing data schemas. The structure model is used to identify rules. The *Define Rules* step defines a rule or an influencing relationship between attributes (e.g., *A* and *B*) in relations for the structure model. The influencing relationship is a relationship between attributes in which there is an unknown causality between the attributes (e.g., *influencing*(*A*, *B*)). If we know the causality, the influencing relationship becomes a causal relationship (e.g., *causal*(*A*, *B*)). For many parent attributes which influence a child attribute (or variable), a brace is used to indicate a set of parent attributes (e.g., *causal*({*A*, *B*}, *C*)). The child attribute is called a *Target Attribute* (or *Variable*). Also, the set of rules should satisfy the No-cycle condition which means that the generated SSBN will contain no directed cycles (Section 2.1).

### 4.2.1 Define Structure Model

The *Define Structure Model* step uses requirements, domain knowledge and/or existing data schemas to develop a structure model. The structure can be represented in a modeling language (e.g., Entity–Relationship (ER) model, Enhanced Entity–Relationship (EER), Relational Model (RM), or Unified Modeling Language (UML)). The structure model can contain information about entities, attributes, and groups for the entities and the attributes (e.g., a relation in RM).

We need to define how to consider the world model in terms of the closed-world assumption and the open-world assumption. The *closed-world assumption* (CWA) means that data, not known to be true, in a database is considered as false, while in the *open-world assumption* (OWA) it is considered as unknown that can be either true or false [Reiter, 1978]. In the world model, entities, relations, and attributes can be treated according to either CWA or OWA. For example, in CWA, if there is a set of disease entities, we assume the only diseases are the ones represented in the RDB. In OWA, there may be other disease entities in addition to the ones represented in the RDB. Considering CWA or OWA depends on the task and the quality of the data or knowledge. If it is sufficient for the task to assume we know all diseases (although in the real world, it is impossible), CWA can be used. As another example, there are a group of trees in a region and we are trying to identify the type of the trees. However, a method to count the trees performs poorly. In this case, it may make sense that data from such a method is treated according to OWA (although we can identify the type of the trees). Therefore, the determination for CWA or OWA for data or knowledge can depend on how these fit well the real world and on the task. This can be an issue of data quality. If our data or knowledge fits well to the real world, we may use CWA. If our data or knowledge does not fit well to the real world, we may use OWA. How to measure such a quality? We may need an approach to qualify the fitness by matching between data and the real world. However, the topic of data quality goes beyond our research.

The original formulation of the relational model assumed a closed world [Date, 2007]. Date [2007] discussed the problem of using OWA in the relational model. Under OWA, data, not known to be true, is considered as *unknown*, which means that we don't know whether it is true or false. Date [2011] discussed that this leads to a three-valued logic (3VL) containing three truth values (e.g., *true*, *false*, and *unknown*). However, the relational model was not developed for such a logic (but it is based on two-valued logic [Date, 2011]). Therefore, query results under the assumption of the three-valued logic for the relation model can be wrong. "*Nulls and 3VL are supposed to be a solution to the "missing information" problem—but I believe I've shown that, to the extent they can be considered a "solution" at all, they're a disastrously bad one.* [Date, 2011]". In this research, we follow a limited closed world assumption to maintain consistency between RM and MEBN in terms of MEBN learning:

**[Assumption 1] No Missing Data:** Values of all RVs for entities explicitly represented in the database are known.

**[Assumption 2] Boolean RV:** For Boolean RVs, if the database does not indicate that the value is true, it is assumed false.

We make no assumptions about entities that have not yet been represented in the database. The purpose of learning is to define a probability distribution for the attributes and relationships for new entities. Relaxing Assumption 1 and Assumption 2 is a topic for future research.

A requirement specifies a query and evidence for the query. The elements of the requirement are used to define corresponding elements in the structure model. For example, suppose that the requirements specify queries for the attributes *Speed* (Query 1.1) and *Speed Report* (Evidence 1.1.1) for a target *g* at a time *t*. Based on these requirements, we know that these two attributes should be included in the structure model. We can then identify additional attributes related to these attributes by expert knowledge. For example, a *TargetType* attribute for the target *g* most likely influences the *Speed* attribute and the *Speed* attribute at the previous time probably influences the current *Speed* attribute. Therefore, these attribute (*TargetType* and *PreviousSpeed*) can be included in the structure model.

In this step, domain knowledge can be used to identify these possible entities, variables, and relationships. Domain knowledge may provide information about possible entities (e.g., time, target, and sensor) involved in the domain situation. For example, Park et al. [2014] suggested possible entities (e.g., the time entity, observer entity, and observed entity). These entities can be a subject which MEBN developers consider for the design of MEBN models in the PSAW domain.

### 4.2.2 Define Rules

In the *Define Rules* step, causal relationships between random variables can be suggested by the PSAW-MEBN reference model. For example, a *Reported Object* RV (e.g., *Speed_RPT*) depends on a *Target Object* RV (e.g., *Speed*). Also, expert knowledge can provide some causal relationships between RVs. For example, an expert can note that the RV *VehicleType* most likely influences the RV *Speed* and an RV *PreviousSpeed* also likely influences the current RV *Speed*. These beliefs from expert knowledge become a causal relationship rule as shown in the following.

    **Rule 1**: *causal*({*VehicleType*, *PreviousSpeed*}, *Speed*)

    **Rule 2**: *causal*(*VehicleType*, *ThreatLevel*)

    **Rule 3**: *causal*({*Speed*, *MTI_Condition*}, *Speed_RPT*)

Rules 1 and 2 are derived from expert knowledge, while Rule 3 is derived from the reference model. Also, in Section 4.2.3.1, the PSAW-MEBN reference model provided knowledge about special context variable types (i.e., context types *ActualObject*, *ObserverOf*, and *Predecessor*) to link entities determined in different MFrags. Thus, the relation *actualobject* is used as the context type *ActualObject*, the relation *observerof* is used as the context type *ObserverOf*, and the relation *predecessor* is used as the context type *Predecessor*.

In the *Define Rules* step, a (conditional) local distribution for an attribute (e.g., the *speed* attribute) can be defined by expert knowledge. In reality, we can meet a situation in which there is no dataset for a rule and all we have is expert knowledge. For example, a conditional local distribution for the *speed* attribute given the RV *VehicleType* can be identified by a domain expert (e.g., if a vehicle type is wheeled, then the speed of the vehicle on a road is normally distributed with a mean of 50MPH and a standard deviation of 20MPH). The rules derived in this step are used in the next step to construct an MTheory and the MTheory will be learned by MEBN parameter learning.

In this step, we determine whether data can be obtained for the attribute, and if so, either collect data or identify an existing dataset. We usually divide the data into a *training* dataset and a *test* dataset. If no data can be obtained, we use the judgment of domain experts to specify the necessary probability distributions. For example, a belief for the target type attribute can be P(*Wheeled*) = 0.8 and P(*Tacked*) = 0.2. If neither data nor expert judgment is available, we consider whether the attribute is really necessary. For this, we can return the *Analyze Requirements* step to modify the requirements.

## 4.3 Construct Reasoning Model

The *Construct Reasoning Model* step develops a reasoning model from a training dataset, a structure model, and rules.

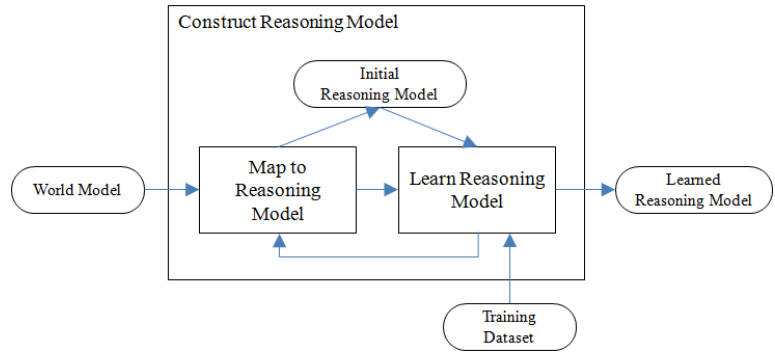

**Fig. 7 Construct Reasoning Model**

This step decomposes into two sub-steps (Fig. 7): (1) a *Map to Reasoning Model* step and (2) a *Learn Reasoning Model* step. The *Map to Reasoning* converts the structure model and rules in the world model to an initial reasoning model. The *Learn Reasoning Model* uses a machine learning method to learn the model from a training dataset.

### 4.3.1 Map to Reasoning Model

In the *Map to Reasoning Model* step, MEBN-RM is used as a reference for a mapping rule between RM and MEBN [Park et al., 2013]. The relations in Fig. 3 can be converted to MFrags in an initial MTheory (MTheory 4.1) using MEBN-RM.

#### 4.3.1.1 Perform Entity-Relationship Normalization

Before performing MEBN-RM [Park et al., 2013], the relations in Table 1 are normalized by the Entity-Relationship Normalization.

**Definition 4.1 (Entity-Relationship Normalization)** Entity-Relationship Normal Form if either its primary key is a single attribute which is not a foreign key, or its primary key contains two or more attributes, all of which are foreign keys.

For example, in the relations in Table 1, we can notice that the relation *VehicleType* has as its primary key a single foreign key imported from the relation *Vehicle*. They (*Vehicle* and *VehicleType*) can be merged into a relation *Vehicle*. The following table shows the normalized table. Note that after the Entity-Relationship Normalization, any foreign key in a relation comes from a certain entity relation (not relationship relation), which has only one attribute for its primary key, so there is no need to indicate which primary key is used for the entity relation and we can simplify the notation for a foreign key (e.g., ***rgn/Region.RID*** and ***t/Time.TID***). For example, the notation of the foreign key for the vehicle (i.e., ***v/Vehicle.VID***) in the relation *Location* (Table 1) can be simplified as ***v/Vehicle***.

**Table 2 Normalized relational dataset from Table 1**

| Time | Region | Vehicle | | Location | | | Situation | | |
|---|---|---|---|---|---|---|---|---|---|
| **TID** | **RID** | **VID** | Vehicle Type | **v/ Vehicle** | **t/Time** | Location /Region | **rgn /Region** | **t /Time** | Threat Level |
| t1 | rgn1 | v1 | Wheeled | v1 | t1 | rgn1 | rgn1 | t1 | High |
| t2 | rgn2 | v2 | Tracked | v1 | t2 | rgn1 | rgn2 | T3 | Low |
| … | … | … | … | … | … | … | … | … | … |

#### 4.3.1.2 Perform MEBN-RM Mapping

The relations in Fig. 3 can be converted to MFrags in an initial MTheory (MTheory 4.1) using MEBN-RM mapping [Park et al., 2013]. For example, the relation *mti_condition* is converted to the MFrag 1, F1, *Mti_Condition* described from the lines 1 and 4. Also, the attributes in relations can be resident nodes in the initial MTheory using the resident node mapping defined in MEBN-RM. For example, the attribute *VehicleType* for the vehicle *v* becomes a resident node VehicleType(*v*). The attribute *Speed* for the vehicle *v* and at the time *t* becomes a resident node Speed(*v*, *t*).

---

**MTheory 4.1**: Initial Threat Assessment

```
1       [F1: MTI_CONDITION
2             [C: IsA (v, VEHICLE), IsA (mti, MTI), IsA (t, TIME)]
3             [R: MTI_Condition(v, mti, t)]
4       ]
5       [F2: VEHICLE
6             [C: IsA (VID, VEHICLE)]
7             [R: VehicleType(VID)]
8       ]
9       [F3: SPEED
10            [C: IsA (v, VEHICLE), IsA (t, TIME)]
11            [R: Speed (v, t)]
12      ]
13      [F4: LOCATION
14            [C: IsA (v, VEHICLE), IsA (t, TIME)]
15            [R: Location (v, t)]
16      ]
```

```
17      [F5: SPEED_REPORT
18              [C: IsA (r, REPORTEDVEHICLE), IsA (t, TIME)]
19              [R: Speed_RPT (r, t)]
20      ]
21      [F6: SITUATION
22              [C: IsA (rgn, REGION), IsA (t, TIME)]
23              [R: ThreatLevel (rgn, t)]
24      ]
25      [F7: REPORTEDVEHICLE
26              [C: IsA (r, REPORTEDVEHICLE)]
27              [R: ActualObject(r)]
28      ]
29      [F8: OBSERVEROF
30              [C: IsA (mti, MTI), IsA (v, VEHICLE)]
31              [R: ObserverOf (mti, v)]
32      ]
33      [F9: PREDECESSOR
34              [C: IsA (pret, TIME), IsA (t, TIME)]
35              [R: Predecessor (pret, t)]
36      ]
```

The initial MTheory, which is directly derived from an RM using MEBN-RM, can be learned using a dataset for each relation associated with the MFrag in the initial MTheory. More specifically, the parameter for the distribution of each RV in the MFrag is learned from a corresponding dataset of the relation for the MFrag. For example, the MFrag *Situation* is derived from the relation *Situation* in Table 2. The parameter for the distribution of the variable *ThreatLevel* (Line 23 in MTheory 4.1) can be learned from the dataset of the attribute *ThreatLevel* (Table 2). An RV (e.g., *ThreatLevel*) in MEBN can contain a *default distribution* which is used for reasoning, for cases in which none of the conditions associated with parent RVs is valid. In MEBN, the parameter for the default distribution should be learned from a dataset containing such cases.

#### 4.3.1.3 Update Reasoning Model using the Rules

The initial MTheory can be updated by the rules defined in Section 4.2.2. We have three rules for the three variables *Speed*, *ThreatLevel*, and *Speed_RPT*. Each variable is associated with its parent variables (e.g., Pa(*ThreatLevel*) = {*VehicleType*})[3]. If the parent variables in the rule for a child variable are in the MFrag where the distribution of the child variable is defined, the dataset for the relation associated with the MFrag is used for learning. For example, suppose that in Table 2, there is an attribute *VehicleSize* in the relation *Vehicle*. The attribute *VehicleSize* becomes a variable *VehicleSize* in the MFrag *Vehicle* using MEBN-RM. If there is a rule such as a *causal*(*VehicleType*, *VehicleSize*), the dataset in the relation *Vehicle* is used to learn the parameter for the distribution of the variable *VehicleSize*. However, if the parent variables in the rule for a child variable are resident in different MFrags where the child variable is not defined, the relations associated with these MFrags for the child variable and the parent variables should be joined to generate a joined dataset containing both datasets for the child and the parents. Then, the dataset for the joined relation from these relations is used for learning. For such a joining, target (or child) variables from a set of rules play an important role. The target variables in the MFrag given their parent variables are learned using the joined relation containing all attributes related to the target variables. In the following, we describe how to join relations in a relational database (e.g., the threat assessment relational database). Rule 1 specifies that the probability distribution of the variable *Speed* depends on the values of variables *PreviousSpeed* and *VehicleType*. To learn the parameter for the variable *Speed* in this situation, it is not enough to use only the dataset from the relation *Speed*, because the dataset doesn't contain information associated with the variable *VehicleType*. Therefore, relations related to each target variable and its parent variables should be joined. For this purpose, we need to define a joining rule. The variable *PreviousSpeed* indicates a variable *Speed* which happens just before a current time, so the relation *Predecessor*, which indicates a previous time and a current time, is also used for this joining for Rule 1. In other words, the relations *Speed*, *VehicleType*, and *Predecessor* are joined.

---
[3] Pa(X) is the set of parent nodes of the node X.

### 4.3.1.3.1  Join Relations

Now, let us discuss how to join these relations. A new dataset from the joined relation is called a *joined dataset*. For example, the attributes (e.g., *VehicleType* and *ThreatLevel*) which are located in different relations can be joined. Joining in RM is an operation to combine two or more relations. In the example, the relation *Vehicle* can be joined to the relation *Situation* through the relation *Location*, because the relation *Location* contains the attributes **_v/Vehicle_**, **_t/Time_**, and Location/Region corresponding to the primary key, **_VID_**, in the relation *Vehicle* and the primary key, **_t/Time_** and **_rgn/Region_**, in the relation *Situation*.

**Table 3 Joined dataset**

| Case | Vehicle.VehicleType | Location.*v* / Vehicle.**VID** | Location.*t* / Situation.*t* | Location.Location / Situation.**rgn** | Situation.ThreatLevel |
|---|---|---|---|---|---|
| 1 | Tracked | Vehicle13 | Time18 | Region6 | High |
| 2 | Tracked | Vehicle15 | Time21 | Region7 | High |
| 3 | Tracked | Vehicle17 | Time24 | Region8 | Low |
| 4 | Tracked | Vehicle19 | Time27 | Region9 | High |
| 5 | Wheeled | Vehicle21 | Time30 | Region10 | High |
| 6 | Wheeled | Vehicle23 | Time33 | Region11 | Low |
| 7 | Wheeled | Vehicle0 | Time2 | Region0 | High |
| 8 | Tracked | Vehicle1 | Time2 | Region0 | High |
| 9 | Tracked | Vehicle2 | Time5 | Region1 | Low |
| 10 | Wheeled | Vehicle3 | Time5 | Region1 | Low |
| 11 | Tracked | Vehicle4 | Time8 | Region2 | High |
| 12 | Tracked | Vehicle5 | Time8 | Region2 | High |
| 13 | Tracked | Vehicle6 | Time11 | Region3 | Low |
| 14 | Tracked | Vehicle7 | Time11 | Region3 | Low |
| 15 | Tracked | Vehicle8 | Time14 | Region4 | High |
| 16 | Tracked | Vehicle9 | Time14 | Region4 | High |
| 17 | Wheeled | Vehicle10 | Time17 | Region5 | High |
| 18 | Wheeled | Vehicle11 | Time17 | Region5 | High |

There are several joining rules (e.g., Cartesian Product, Outer Join, Inner Join, and Natural Join) [Date, 2011]. Table 3 shows an illustrative example of a joined dataset derived from Table 2 using Inner Join. Inner Join produces all tuples from relations as long as there is a match between values in the columns being joined. Table 3 shows the result of performing an inner join of the relations *Situation* and *Vehicle* through the relation *Location* and then selecting the columns to be used for learning. The rows (or tuples) in the relations *Situation* and *Vehicle* are joined when rows of the attributes **_v/Vehicle_**, **_t/Time_**, and Location/Region in the relation *Location* match rows of the attribute **_VID_** in the relation *Vehicle* and rows of the attributes **_rgn/Region_** and **_t/Time_** in the relation *Situation*. The first column denotes cases for the matched rows. The second column (Vehicle.VehicleType) denotes the rows from the attribute *VehicleType* of the relation *Vehicle* in Table 2. The third column (Location.*v* and Vehicle.**_VID_**) denotes the matched rows between the attribute ***v*** from the relation *Location* and the attribute **_VID_** from the relation *Vehicle*. The fourth column (Location.***t*** and Situation.***t***) denotes the matched rows between the attribute ***t*** from the relation *Location* and the attribute ***t*** from the relation *Situation*. The fifth column (Location.<u>Location</u> and Situation.***rgn***) denotes the matched rows between the attribute <u>Location</u> from the relation *Location* and the attribute ***rgn*** from the relation *Situation*. The sixth column (Situation.ThreatLevel) denotes the rows from the attribute *ThreatLevel* from the relation *Situation*.

Table 3 shows the joined dataset for the attributes *VehicleType* and *ThreatLevel*. Now, let us assume that the attribute *ThreatLevel* will be a target variable depending on the variable *VehicleType* (i.e., Rule 2: *causal*(*VehicleType*, *ThreatLevel*)). For each instance of the target variable *ThreatLevel*, Table 3 provides relevant information about all the configurations of its parents (i.e., the parent variable *VehicleType*). For example, there is the value *High* for the Threat level in the situation at *Region5* in *Time17* (i.e., Cases 17 and 18). The value *High* is associated with the wheeled *Vehicle10* and the wheeled *Vehicle11*. In other words, two parent instances (i.e., the wheeled *Vehicle10* and the wheeled *Vehicle11*) influence the target instance (i.e., the value *High*). The following shows a query script[4] which is an example using Inner Join for Table 3.

**SQL script 4.1**: Joining for Table 3

```
1   SELECT
2       Vehicletype, Location.v, Location.t, Location.Location, ThreatLevel
3   FROM Situation
4
5   JOIN Location ON
6       Situation.rgn = Location.Location &&
7       Situation.t = Location.t
8
9   JOIN Vehicle ON
10      Vehicle.VID = Location.v
```

SQL script 4.1 joins the relations *Situation* and *VehicleType* through the relation *Location*. In other words, the rows (or tuples) in the relations *Situation* and *VehicleType* are joined as shown Table 3 in which the two attributes (*VehicleType*, *ThreatLevel*) are connected through the attributes of the relation *Location*. The joined table shows how the dataset of the attribute *VehicleType* and the dataset of the attribute *ThreatLevel* are linked.

We introduced how to join relations according to given rules. In the following, we discuss how to update an MFrag from the given rules. The initial threat assessment MTheory (MTheory 4.1) was constructed by MEBN-RM. Each MFrag in the initial MTheory contains resident nodes without any causal relationship between the resident nodes. The given rules enable the resident nodes to specify such causal relationships. Therefore, the MFrag in the initial MTheory may be changed according to the updated resident nodes with the causal relationships by the given rules. This process contains three steps: Construct input/parent nodes, Construct context nodes, and Refine context nodes.

### *4.3.1.3.2 Construct Input/Parent Nodes*

A rule denotes a target variable and its parent variables. The joined table for such a given rule contains parents of the resident node (i.e., the target variable) that may be resident in another MFrag and need to be added as input nodes for the resident node. For example, we defined a set of rules in the *Define World Model* step (e.g., Rule 2: *causal*(*VehicleType*, *ThreatLevel*)). In

---
[4] In this research, we used MySQL, an open-source relational database management system, and Structured Query Language (SQL) supported by MySQL.

MTheory 4.1, for the target variable *ThreatLevel* in the MFrag *Situation*, its parent *VehicleType* is defined in the MFrag *Vehicle*. The parent variable *VehicleType* should be an input node in the MFrag *Situation*. The following MFrag shows the updated result for the MFrag *Situation* using Rule 2.

| | |
|---|---|
| **MFrag 4.1**: Situation | |
| 1 | [C: IsA (*rgn*, REGION), IsA (*t*, TIME)] |
| 2 | [C: IsA (*VID*, VEHICLE)] |
| 3 | [R: ThreatLevel (*rgn*, *t*) |
| 4 | [IP: VehicleType (*VID*)] |
| 5 | ] |

The primary key for *VehicleType* is *VID* associated with the entity VEHICLE, so IsA (*v*, VEHICLE) is added in the updated MFrag *Situation* (MFrag 4.1).

### 4.3.1.3.3 Construct Context Nodes

In this step, additional context nodes (other than *IsA* context nodes) are added to the updated MFrag. For this, we can use a joining script (e.g., SQL script 4.1) used for joining relations. In SQL script 4.1, there are conditions for joining. (e.g., *Situation.rgn = Location.Location*, *Situation.t = Location.t*, and *Vehicle.VID = Location.v*). These conditions are represented as context nodes. For example, the condition *Situation.rgn = Location.Location* can be a context node *rgn* = Location(*v*, *t1*), where the ordinary variable *rgn* comes from the primary key *rgn* in the relation *Situation*, the first *v* comes from the relation *Location*, and the second *t1* comes from the relation *Location*. Note that although the primary key *t* for the attribute *ThreatLevel* and the primary key *t* for the attribute *Location* are same, they must be given different ordinary variable names in the context nodes, because they refer to different entities. For example, Location(*v*, *t*) for the attribute *Location* can be changed to Location(*v*, *t1*). The condition *Situation.t = Location.t* can be a context node *t = t1*, where the first *t* comes from the relation *Situation* associated with the attribute *ThreatLevel* and the second *t1* comes from the relation *Location* associated with the attribute *Location*. From the above process, the following script can be developed.

| | |
|---|---|
| **MFrag 4.2**: Situation | |
| 1 | [C: IsA (*rgn*, REGION), IsA (*t*, TIME)] |
| 2 | [C: IsA (*VID*, VEHICLE)] |
| 3 | [C: IsA (*v*, VEHICLE), IsA (*t1*, TIME)] |
| 4 | [C: *rgn* = Location (*v*, *t1*)] |
| 5 | [C: *t = t1*] |
| 6 | [C: *VID = v*] |
| 7 | [R: ThreatLevel (*rgn*, *t*) |
| 8 | [IP: VehicleType (*VID*)] |
| 9 | ] |

The primary key for the attribute *Location* are *v* and *t*, so the *IsA* context nodes IsA (*v*, VEHICLE) and IsA (*t1*, TIME) are added to MFrag 4.2.

### 4.3.1.3.4 Refine Context Nodes

In MFrag 4.2, we notice that two equal-context nodes (i.e., [C: *t = t1*] in Line 5 and [C: *VID = v*] in Line 6) indicate conditions that entities must be equal. Consequently, the equal-context node indicates that they are the same entity. The above script can be simplified by removing ordinary variables sharing the same entity and equal-context nodes as shown MFrag 4.2.

**MFrag 4.3**: Situation
| | |
|---|---|
| 1 | [C: IsA (*v*, VEHICLE)] |
| 2 | [C: IsA (*t*, TIME), IsA (*rgn*, REGION)] |
| 3 | [C: *rgn* = Location (*v*, *t*)] |
| 4 | [R: ThreatLevel (*rgn*, *t*) |
| 5 | [IP: VehicleType (*v*)] |
| 6 | ] |

The *Learn Reasoning Model* step applies MTheory learning from relational data. In this research, we focus on MEBN parameter learning given a training dataset *D* in RM and an initial MTheory *M*. Before introducing MEBN parameter learning, some definitions are introduced in the following subsections.

### 4.3.2 Definitions for Class Local Distribution and Instance Local Distribution

We introduced Definition 2.2 (MFrag), Definition 2.3 (MNode), and Definition 2.4 (MTheory) for MEBN in Section 2. An MTheory is composed of a set of MFrags *F* on the MTheory (i.e., $M = \{F_1, F_2, ... , F_n\}$) conditions (e.g., no-cycle, bounded causal depth, unique home MFrags, and recursive specification condition [Laskey, 2008]) in Section 2. An MFrag *F* is composed of a set of MNodes *N* and a graph *G* for *N* (i.e., $F = \{N, G\}$). An MNode is composed of a function or predicate of FOL *ff* and a class local distribution (*L*) (i.e., $N = \{ff, L\}$).

A CLD specifies how to define local distributions for instantiations of the MNode. The following CLD 4.1 and ILD 4.1 show illustrative examples for a CLD (Class Local Distribution) and an ILD (Instance Local Distribution), respectively (recall that these examples were discussed in Section 2). CLD 4.1 defines a distribution for the threat level in a region. If there are no tracked vehicles, the *default probability distribution* described in Line 6 is used. The default probability distribution in a CLD is used for ILDs generated from the CLD, when no nodes meet the conditions defined in the MFrag for parent nodes.

This CLD is composed of a *class parent condition $CPC_i$* and a *class-sub-local distribution $CSD_i$*. A CPC indicates a condition whether a CSD associated with the CPC is valid. The CSD (class-sub-local distribution) is a sub-probability distribution which specifies how to define a local distribution under a condition in an RV derived from an MNode. For example, the first line in CLD 4.1 is $CPC_1$ which indicates a condition of the first class-sub-local distribution $CSD_1$. In this case, the condition means that "if there is an object whose type is *Tracked*". If this is satisfied (i.e., $CPC_1$ is valid), then $CSD_1$ is used. A CPC can be used for a default probability distribution. In such a case, it is called a *default CPC* specified by $CPC_d$ and also the CSD associated with $CPC_d$ is called a default CSD, $CSD_d$.

**CLD 4.1 [Discrete CLD]:** ThreatLevel(*rgn*, *t*)
| | | |
|---|---|---|
| 1 | $CPC_1$: | **if** some *v* **have** (VehicleType = *Tracked*) [ |
| 2 | $CSD_1$: | $High = \Theta_{1.1}, Low = \Theta_{1.2}$ |
| 3 | $CPC_2$: | ] **else if** some *v* **have** (VehicleType = *Wheeled*) [ |
| 4 | $CSD_2$: | $High = \Theta_{2.1}, Low = \Theta_{2.2}$ |
| 5 | $CPC_d$: | ] **else** [ |
| 6 | $CSD_d$: | $High = \Theta_{d.1}, Low = \Theta_{d.2}$  ] |

For this case, we assume that the MNode contains two states (*High* and *Low*) and the discrete parent the RV *VehicleType*(*v*) has two states (*Tracked* and *Wheeled*). The pair of $CSD_1$ and $CSD_1$ (in Line 1 and 2) is for *VehicleType*(*v*) = *Tracked*. The pair of $CSD_2$ and $CSD_2$ (in Line 3 and 4) is for VehicleType(*v*) = *Wheeled*. The pair of $CPC_d$ and $CSD_d$ (in Line 5 and 6) is for a default distribution.

The following ILD 4.1 shows the ILD derived from the above CLD given one region entity *region1* and one vehicle entity *v1*. Like the CLD, the ILD is composed of an *instance parent condition $IPC_i$* and an *instance-sub-local distribution $ISD_i$*. The IPC indicates a condition whether the ISD associated with the IPC is valid. The ISD is a probability distribution which is defined in an ILD of a random variable.

---

**ILD 4.1:** ILD with one region and one vehicle

```
1            P(ThreatLevel_region1 | VehicleType_v1 )
2   IPC₁:    if( VehicleType_V1 == Wheeled ) {
3   ISD₁:         High = Θ₁.₁; Low = Θ₁.₂;
4   IPC₂:    } else if( VehicleType_V1 == Wheeled ) {
5   ISD₂:         High = Θ₂.₁; Low = Θ₁.₂;
6            }
```
---

Now, consider a situation in which there is a region containing no vehicles. In this case, the default probability distribution in CLD 4.1 is used for such an ILD (i.e., ILD 4.2), because all conditions associated with parent nodes (i.e., $CPC_1$ and $CPC_2$ in CLD 4.1) are not valid.

---

**ILD 4.2:** Default ILD with one region without any vehicle

```
1            P(ThreatLevel_region1)  =
2   IPC₁:    {
3   ISD₁:         High = Θ_d.1; Low = Θ_d.2;
4            }
```
---

Now, we introduce the ILD formally.

**Definition 4.2 (Instance Local Distribution)** An *instance local distribution $L^I$* for a random variable *rv* in a Bayesian network (Definition 2.1) is a function defining the probability distribution for the random variable *rv*. It consists of a set of pairs ($IPC_i$, $ISD_i$) of an *instance parent condition $IPC_i$* and an *instance-sub-local distribution $ISD_i$*, and a rule for mapping an *instance parent condition $IPC_i$* into an *instance-sub-local distribution $ISD_i$*.

An ILD is derived from a CLD given entity information. For example, ILD 4.1 in the above example is derived from CLD 4.1 given the three vehicle entities. Once an ILD is derived from a CLD, the ILD contains a set of pairs ($IPC_i$, $ISD_i$). In the following, the CLD is introduced formally.

**Definition 4.3 (Class Local Distribution)** A *class local distribution* (CLD) $L^C$ (or simply *L*) for an MNode (Definition 2.3) is a function defining uncertainty for the MNode. It consists of a set of pairs ($CPC_i$, $CSD_i$) of a *class parent condition $CPC_i$* and a *class-sub-local distribution $CSD_i$*, and a rule for mapping it ($CPC_i$, $CSD_i$) into an *instance local distribution* (ILD) $L^I$.

A class local distribution defines a general rule for specifying distributions for instantiations of its random variables for specific entities. A CLD can refer to a parameterized family of distributions (e.g., normal distribution, categorical distribution). In this case, the CLD definition includes a specification of the parameters. For example, a class local distribution $CLD_1$ can represent a set of normal distributions for CSDs in $CLD_1$ and this $CLD_1$ can be called a *normal distribution CLD* (i.e., TYPE($CLD_1$) = Normal Distribution CLD). A CSD can contain a set of parameters for its distribution. For example, $CSD_1$ in CLD 4.1 is a distribution containing two parameters $\theta_{1.1}$ and $\theta_{1.2}$. We can think of a parameter function returning a set of parameters from a CSD (i.e., $\Theta(CSD_1) = \{\theta_{1.1}, \theta_{1.2}\}$).

CLDs may be discrete or continuous. According to combination of the CLD types and parent CLD types, there are six categories for a CLD: (1) a discrete CLD with discrete parents, (2) a discrete CLD with continuous parents, (3) a discrete CLD with both discrete and continuous parents, (4) a continuous CLD with discrete parents, (5) a continuous CLD with continuous parents, and (6) a continuous CLD with both discrete and continuous parents.

When a node has discrete parent nodes, *influence counts* (IC), the number of distinct entities in CPC, can be used to define a CLD. For example, we can think of a CLD for the MNode *ThreatLevel*(*rgn*, *t*) described by LPDL as shown the following. CLD 4.2 is the case of a discrete CLD with discrete parents.

| | | |
|---|---|---|
| **CLD 4.2 [Inverse Cardinality Average]:** ThreatLevel (*rgn*, *t*) | | |
| 1 | $CPC_1$: | **if** some *v* **have** (VehicleType = *Tracked* ) [ |
| 2 | $CSD_1$: | *High* = 1 - $\Theta$/(CARDINALITY(*v*) + 1), *Low* = 1 - *High* |
| 3 | $CPC_d$: | ] **else** [ |
| 4 | $CSD_d$: | *High* = 0.1, *Low* = 0.9 |
| 5 | | ] |

We name CLD 4.2 an *Inverse Cardinality Average*. Thus, the type of the class local distribution is the inverse cardinality average (i.e., TYPE(CLD 4.2) = Inverse Cardinality Average CLD). CLD 4.2 consists of two CSDs ($CSD_1$ and $CSD_d$). $CSD_1$ contains a parameter $\theta$, where $0 < \theta < 1$, as shown CLD 4.2. CLD 4.2 represents probabilistic knowledge of how the threat level of a region is measured depending on the vehicle type of detected objects. For example, if in a region there are many tracked vehicles (e.g., Tanks), the threat level of the region at a certain time will be high. The influence counting (IC) function CARDINALITY(obj) returns the number of tracked vehicles from parents nodes. If there are many tracked vehicles, the probability of the state *High* increases. If there is no tracked vehicles, the default probability distribution (i.e., $CSD_d$) described in Line 4 is used for the CLD of the MNode ThreatLevel(*rgn*, *t*). Thus, it indicates a situation in peace time.

Here is another CLD example. CLD 4.3 shows the case of the continuous CLD with hybrid parents. For this case, we assume that there is an MNode Range(*v*, *t*) which is a parent node of the MNode ThreatLevel(*rgn*, *t*) and means a range between the region *rgn* and the vehicle *v* at a time *t*.

| | | |
|---|---|---|
| **CLD 4.3 [Hybrid Cardinality]:** ThreatLevel(*rgn*, *t*) | | |
| 1 | $CPC_1$: | **if** some *v* **have** (VehicleType = *Tracked* ) [ |
| 2 | $CSD_1$: | CARDINALITY(*v*) / average( *Range* ) + NormalDist($\Theta$, 5) |
| 3 | $CPC_d$: | ] **else** [ |
| 4 | $CSD_d$: | NormalDist(10, 5) |
| 5 | | ] |

The meaning of CLD 4.3 is that the threat level in the region is the number of tracked vehicles divided by an average of the ranges of vehicles and then plus a normally distributed error with a mean of $\Theta$ and a variance of 5. If there is no tracked vehicles, the default probability distribution, NormalDist(10, 5), described in Lines 4 is used. If there are continuous parents, various numerical *aggregating* (AG) functions (e.g., *average*, *sum*, and *multiply*) can be used. For example, if there are three continuous parents *Range1*, *Range2*, and *Range3*, the numerical aggregating functions average, sum, and multiply will construct three IPDs $IPD_1 = (Range1 + Range2 + Range3)/3$, $IPD_2 = (Range1 + Range2 + Range3)$, and $IPD_3 = (Range1 * Range2 * Range3)$, respectively.

The above CLDs 4.2 and 4.3 are based on an influence counting (IC) function for discrete parents and an aggregating (AG) function for continuous parents. Using such a function is related to the aggregating influence problem, which treats many instances from a parent RV.

The CLD 4.1 uses a very simple aggregation rule that treats all counts greater than zero as equivalent. In other words, a shared parameter in a CSD is learned from all instances of the parent RV with counts greater than zero. For example, with CLD 4.1, suppose that there are two cases: In Case 1, there is one tracked vehicle. And in Case 2, there are two tracked vehicles. For Case 1, one *VehicleType* RV is constructed and $CSD_1$ (Line 1) in CLD 4.1 is used for the parameter of the distribution for the *ThreatLevel*. For Case 2, two *VehicleType* RVs are constructed and also $CSD_1$ (Line 1) in CLD 4.1 is used for the parameter of the distribution for the *ThreatLevel*, although there are two tracked vehicles. Thus, the shared parameter (i.e., $High = \Theta_{1.1}$ and $Low = \Theta_{1.2}$) for $CSD_1$ in CLD 4.1 is used regardless of the number of the parent instances (i.e., one vehicle in Case 2, two vehicles in Case 2, and so on). In the following sections, we use such a simple aggregation rule for MEBN parameter learning.

### 4.3.3 Dataset for Class-Sub-Local Distribution (CSD)

A CLD can contain class parent conditions (CPC). Each CPC requires its own dataset to be learned to a class-sub-local distribution CSD associated with the CPC. For example, CLD 4.1 contains three CPCs ($CPC_1$, $CPC_2$, and $CPC_d$). Each CPC requires its own dataset. Such a dataset can be classified by two categories: (1) A dataset for a *common CPC* (e.g., $CPC_1$ and $CPC_2$) and (2) a dataset for a *default CPC* (e.g., $CPC_d$). In this section, we introduce how to get the dataset for a common CPC first. Then we present how to get the dataset for a default CPC.

**Table 4 CSD dataset**

| CPC | Case | Vehicle.VehicleType | Location.$v$ Vehicle.***VID*** | Location.$t$ Situation.$t$ | Location.Location Situation.***rgn*** | Situation.ThreatLevel |
|---|---|---|---|---|---|---|
| $CPC_1$ ($GC_1$) | 1 | Tracked | Vehicle13 | Time18 | Region6 | High |
| | 2 | Tracked | Vehicle15 | Time21 | Region7 | High |
| | 3 | Tracked | Vehicle17 | Time24 | Region8 | Low |
| | 4 | Tracked | Vehicle19 | Time27 | Region9 | High |
| | 8 | Tracked | Vehicle1 | Time2 | Region0 | High |
| | 9 | Tracked | Vehicle2 | Time5 | Region1 | Low |
| | 11 | Tracked | Vehicle4 | Time8 | Region2 | High |
| | 12 | Tracked | Vehicle5 | Time8 | Region2 | High |
| | 13 | Tracked | Vehicle6 | Time11 | Region3 | Low |
| | 14 | Tracked | Vehicle7 | Time11 | Region3 | Low |
| | 15 | Tracked | Vehicle8 | Time14 | Region4 | High |
| | 16 | Tracked | Vehicle9 | Time14 | Region4 | High |
| $CPC_2$ ($GC_2$) | 5 | Wheeled | Vehicle21 | Time30 | Region10 | High |
| | 6 | Wheeled | Vehicle23 | Time33 | Region11 | Low |
| | 7 | Wheeled | Vehicle0 | Time2 | Region0 | High |
| | 10 | Wheeled | Vehicle3 | Time5 | Region1 | Low |
| | 17 | Wheeled | Vehicle10 | Time17 | Region5 | High |
| | 18 | Wheeled | Vehicle11 | Time17 | Region5 | High |

Table 3 is a joined dataset for the common CPC (i.e., $CPC_1$ and $CPC_2$). It can be sorted according to each CPC as shown in Table 4. For example, the $CPC_1$ in CLD 4.1 defines that it is only valid if a case contains a tracked vehicle. Therefore, by $CPC_1$, we can sort the joined dataset in Table 3. Thus, the cases 1, 2, 3, 4, 8, 9, 11, 12, 13, 14, 15, and 16 are selected for $CSD_1$, while other cases are used for $CSD_2$ (Table 4). We call this dataset a *CSD dataset*.

**Definition 4.4 (CSD Dataset)** Let there be a dataset $D = \{C_1, C_2, \ldots, C_n\}$, where $C_i$ is each case (or row), and a CLD $L = \{(CPC_1, CSD_1), (CPC_2, CSD_2), \ldots, (CPC_m, CSD_m)\}$. A *CSD Dataset* (*CD*) is a dataset which is grouped by matching each class parent condition $CPC_j$ of *L* and each case $C_i$ in *D*. The set of grouped cases $GC_j = \{C_1, C_2, \ldots, C_l\}$ is assigned to a corresponding class parent condition $CPC_j$.

For an RV, if there are cases for which the conditions associated with the parent RVs are not satisfied, the dataset for a default CPC is required. The dataset for the default CPC (i.e., $CPC_d$) can be obtained by excluding the joined dataset from the original dataset. This is necessary because we need a dataset which doesn't include cases for which the conditions associated with the parent RVs are satisfied. For example, in Table 2, there is an original dataset for the *ThreatLevel* RV (i.e., the dataset in the relation *Situation*). Table 3 shows a joined dataset associated with $CPC_1$ and $CPC_2$. The dataset for the default $CPC_d$ can be derived by subtracting the joined dataset (Table 3) from the original dataset (Table 2). For example, the following is a SQL script to extract the default dataset for the *ThreatLevel* RV from the joined dataset.

**SQL script 4.2**: SQL script for the default dataset of the *ThreatLevel* RV

```
1   SELECT
2       Situation.rgn, Situation. t, Situation .ThreatLevel
3   FROM Situation
4   WHERE NOT EXISTS (
5   SELECT *
6   FROM Location, Vehicle
7   WHERE
8       Situation.rgn  =  Location.Location &&
9       Situation.t  =  Location.t &&
10      Vehicle.VID = Location.v
11  )
```

The dataset for the *ThreatLevel* RV comes from the relation *Situation* (Line 3). When the dataset is selected, there is a condition (Line 4) in which the dataset should not include a joined dataset derived by Line 5~11. Using this script, the default dataset for the *ThreatLevel* RV is obtained and means the threat level at a certain region, where there is no vehicle.

In the following subsections, a training dataset *D* means the CSD dataset for a certain CLD.

### 4.3.4   Parameter Learning

In this section, we introduce a parameter learning method to estimate parameters of a class local distribution *L* given a training dataset *D* (i.e., a CSD dataset). We can think of a basic type of CLD for a discrete case and a continuous case. For the discrete case, Dirichlet distribution can be used (Section 4.3.4.1), while for the continuous case, Conditional Gaussian distribution can be used (Section 4.3.4.2). We introduce parameter learning for these types. In Definition 2.2 (MNode), a predicate RV for MEBN was discussed. Learning the parameter of the distribution for such a predicate RV, corresponding to a Boolean RV with possible values *true* and *false*, from a relational database is discussed in Section 4.3.4.3.

### 4.3.4.1   Dirichlet Distribution Parameter Learning

Details on Dirichlet distribution parameter learning can be found in Appendix A. Dirichlet

distribution is commonly used because it is conjugate to the multinomial distribution. With a Dirichlet prior distribution, the posterior predictive distribution has a simple form [Heckerman et al., 1995][Koller & Friedman, 2009].

As an illustrative example of the Dirichlet distribution parameter learning for a CLD, we use CLD 4.1. Parameter learning for this CLD is to estimate $CSD_1$'s parameters ($\Theta_{1.1}$ and $\Theta_{1.2}$), and $CSD_2$'s parameters ($\Theta_{2.1}$ and $\Theta_{2.2}$), and $CSD_d$'s parameters ($\Theta_{d.1}$ and $\Theta_{d.2}$). To estimate these parameters, we can use the following predictive distribution using a Dirichlet conjugate prior, discussed in Appendix A. Equation 4.1 shows the posterior predictive distribution for the value $x_k$ of the RV $X$ given a parent value $a$, the dataset $D$, and a hyperparameter α for the Dirichlet conjugate prior.

$$P(X = x_k \mid A = a, D, \alpha) = \frac{\alpha_{x_k|a} + C[x_k, a]}{\sum_{q=1}^{N}(\alpha_{x_q|a} + C[x_q, a])}, \tag{4.1}$$

where a value $x_k \in Val(X)$, $a \in Val(Pa(X) = A)$, $C[x_q, a]$ is the number of times outcome $x_q$ in $X$ and its parent outcome $a$ in $A$ appears in $D$, a hyperparameter $\alpha = \{\alpha_{x_1|a}, ..., \alpha_{x_N|a}\}$, and $N = |Val(X)|$.

For the case of the $CPC_1$ and $CSD_1$, we can use the set of grouped cases $GC_1$ in Table 4 as a training dataset. And $CSD_1$ has two parameters $\Theta_{1.1}$ (for *High*) and $\Theta_{1.2}$ (for *Low*). For the parameters $\Theta_{1.1}$, we can use Equation 4.1 such as $\Theta_{1.1}$ = P(ThreatLevel = *High* | VehicleType = *Tracked*, D = $GC_1$, α), where α = {$\alpha_{High|Tracked}$, $\alpha_{Low|Tracked}$}. If there were previously one case for *High|Tracked* and two cases *Low|Tracked*, $\alpha_{High|Tracked}$ = 1 and $\alpha_{Low|Tracked}$ = 2 are used. This approach uses again for the case of the $CPC_2$ and $CSD_2$. To learn the parameter for the $CSD_d$, the default dataset discussed in Section 4.3.3 is required. The parameter $\Theta_{d.1}$ and $\Theta_{d.2}$ can be learned from the default dataset using Equation 4.1 as the case of the $CPC_1$ and $CSD_1$.

### 4.3.4.2 Conditional Linear Gaussian Distribution Parameter Learning

Parameters for conditional Gaussian distribution can be estimated using multiple-regression. In this section, we introduce parameter learning of a conditional linear Gaussian CLD using linear regression. The following CLD shows an illustrative example of a conditional linear Gaussian CLD for the RV Speed_RPT($r$, $t$). The CLD of the RV is a continuous CLD with hybrid parents (*MTI_Condition* and *Speed*). In this case, we assume that the discrete parent RV MTI_Condition($v$, $mti$, $t$) has two states (*Good* and *Bad*) and the RV Speed($v$, $t$) is continuous.

| | | |
|---|---|---|
| **CLD 4.4 [Conditional Linear Gaussian]:** Speed_RPT($r$, $t$) | | |
| 1 | $CPC_1$: | **if** some *v.mti.t* **have** (MTI_ Condtion = *Good*) [ |
| 2 | $CSD_1$: | $\Theta_{1.0} + \Theta_{1.1}$* Speed + NormalDist(0, $\Theta_{1.2}$) |
| 3 | $CPC_2$: | **if** some *v.mti.t* **have** (MTI_ Condtion = *Bad*) [ |
| 4 | $CSD_2$: | $\Theta_{2.0} + \Theta_{2.1}$* Speed + NormalDist(0, $\Theta_{2.2}$) |
| 5 | $CPC_d$: | ] **else** [ |
| 6 | $CSD_d$: | $\Theta_{d.0}$ + NormalDist(0, $\Theta_{d.2}$) |
| 7 | | ] |

Parameter learning for this CLD is to estimate $CSD_1$'s parameters ($\Theta_{1.0}$, $\Theta_{1.1}$ and $\Theta_{1.2}$), $CSD_2$'s parameters ($\Theta_{2.0}$, $\Theta_{2.1}$ and $\Theta_{2.2}$), and $CSD_d$'s parameters ($\Theta_{d.0}$, $\Theta_{d.1}$ and $\Theta_{d.2}$). We can write this situation more formally. If $X$ is a continuous node with $n$ continuous parents $U_1$, …, $U_n$ and $m$ discrete parents $A_1$, …, $A_m$, then the conditional distribution p($X$ | $u$, $a$) given parent states $U = u$

and $A = a$ has the following form:

$$p(X \mid \boldsymbol{u}, \boldsymbol{a}) = \mathcal{N}\big(\mathrm{L}^{(\boldsymbol{a})}(\boldsymbol{u}), \sigma^{(\boldsymbol{a})}\big), \tag{4.2}$$

where $\mathrm{L}^{(\boldsymbol{a})}(\boldsymbol{u}) = m^{(\boldsymbol{a})} + b_1^{(\boldsymbol{a})} u_1 + \cdots + b_n^{(\boldsymbol{a})} u_n$ is a linear function of the continuous parents, with intercept $m^{(\boldsymbol{a})}$, coefficients $b_i^{(\boldsymbol{a})}$, and standard deviation $\sigma^{(\boldsymbol{a})}$ that depends on the state $\boldsymbol{a}$ of the discrete parents. Given $\mathrm{CPC}_j$ (i.e., given the state $\boldsymbol{a}_j$), estimating the parameters the intercept $m^{(\boldsymbol{a}_j)}$, coefficients $b_i^{(\boldsymbol{a}_j)}$, and standard deviation $\sigma^{(\boldsymbol{a}_j)}$ corresponds to estimating the CSD's parameters $\Theta_{j,0}$, $\Theta_{j,1}$ and $\Theta_{j,2}$, respectively.

The following shows multiple linear regression which is modified from [Rencher, 2003]. $\mathrm{L}^{(\boldsymbol{a})}(\boldsymbol{u})$ can be rewritten, if we suppose that there are $k$ observations (Note that for one CSD case, we can omit the state $\boldsymbol{a}$, because we know it).

$$\mathrm{L}_i(\boldsymbol{u}) = m + b_1 u_{i1} + \cdots + b_n u_{in} + \sigma_i, i = 1, \ldots, k \tag{4.3}$$

where $i$ indexes the observations. For convenience, we can write the above equation more compactly using matrix notation:

$$\mathbf{l} = \boldsymbol{U}\boldsymbol{b} + \boldsymbol{\sigma}, \tag{4.4}$$

where $\mathbf{l}$ denotes a vector of instances for the observations, $\boldsymbol{U}$ denotes a matrix containing all continuous parents in the observations, $\boldsymbol{b}$ denotes a vector containing an intercept $m$ and a set of coefficients $b_i$, and $\boldsymbol{\sigma}$ denotes a vector of regression residuals. The following equations show these variables in forms of vectors and a matrix.

$$\mathbf{l} = \begin{bmatrix} \mathrm{L}_1(\boldsymbol{u}) \\ \mathrm{L}_2(\boldsymbol{u}) \\ \ldots \\ \mathrm{L}_k(\boldsymbol{u}) \end{bmatrix} \quad \boldsymbol{U} = \begin{bmatrix} 1 & u_{11} & \ldots & u_{1n} \\ 1 & u_{21} & \ldots & u_{2n} \\ \ldots & \ldots & \ldots & \ldots \\ 1 & u_{k1} & \ldots & u_{kn} \end{bmatrix} \quad \boldsymbol{b} = \begin{bmatrix} m \\ b_1 \\ \ldots \\ b_k \end{bmatrix} \quad \boldsymbol{\sigma} = \begin{bmatrix} \sigma_1 \\ \sigma_2 \\ \ldots \\ \sigma_k \end{bmatrix} \tag{4.5}$$

From the above settings, we can derive an optimal vector for the intercept and the set of coefficients $\widehat{\boldsymbol{b}}$.

$$\widehat{\boldsymbol{b}} = (\boldsymbol{U}^T \boldsymbol{U})^{-1} \boldsymbol{U}^T \mathbf{l} \tag{4.6}$$

Also, we can derive the optimal standard deviation $\widehat{\sigma}$ from the above linear algebra term [Rencher, 2003].

$$\hat{\sigma} = \sqrt{\frac{(l - U\hat{b})^T(l - U\hat{b})}{k - n - 1}} \tag{4.7}$$

Using the above equations, the optimal parameters can be estimated. For $CPC_1$ in CLD 4.4, $CSD_1$ can be the following.

$$p(X \mid \text{Speed}, \text{MTI\_Condtion} = Good) = \mathcal{N}\left(\theta_{1.0} + \text{Speed} * \theta_{1.1}{}^{(Good)}, \theta_{1.2}{}^{(Good)}\right).$$

In this section, we discussed how to learn parameters for the conditional linear Gaussian CLD using linear regression. For a conditional nonlinear Gaussian CLD, we can use nonlinear regression. In this section, we didn't consider incremental parameter learning for the conditional linear Gaussian CLD. For this, we can Bayesian regression [Press, 2003], which is more robust to overfitting than the traditional multiple-regression.

### 4.3.4.3  Parameter Learning for the Distribution of the Predicate/Boolean RV

The parameter of the distribution for a predicate or Boolean RV (Definition 2.2) can be learned from a relational database. To introduce predicate RV parameter learning, the following relations in Table 5 as an illustrative example are used to learn the parameter of the distribution for a predicate RV *Communicate*. The following table contains three relations *Vehicle*, *Communicate*, and *Meet*. The relation *Communicate* means that two vehicles communicate with each other by exchanging radio waves. The relation *Meet* means that two vehicles meet each other by locating in close proximity to each other.

**Table 5 Communicate Relation and Meet Relation**

| Vehicle | Communicate | | Meet | |
|---|---|---|---|---|
| *VID* | *VID1/Vehicle* | *VID2/Vehicle* | *VID1/Vehicle* | *VID2/Vehicle* |
| v1 | v1 | v2 | v1 | v2 |
| v2 | v2 | v3 | v2 | v3 |
| v3 | v3 | v4 | v1 | v4 |
| v4 | | | | |

The above relationship relations (i.e., *Communicate* and *Meet*) show true cases for predicates. For example the relation *Communicate* contains the true cases {{v1, v2}, {v2, v3}, {v3, v4}}. However, relationship relations do not explicitly represent false cases for the predicates. By converting the above relations to the following relations, we can see the false cases explicitly. This conversion is justified by CWA. Thus, if a case in a relationship relation is not true, it is assumed to be false.

**Table 6 Converted Relations from Communicate Relation and Meet Relation**

| Vehicle | | Communicate | | | | Meet | |
|---|---|---|---|---|---|---|---|
| *VID* | | *VID1/Vehicle* | *VID2/Vehicle* | Communicate | *VID1/Vehicle* | *VID2/Vehicle* | Meet |
| v1 | | v1 | v2 | True | v1 | v2 | True |
| v2 | | v1 | v3 | False | v1 | v3 | False |
| v3 | | v1 | v4 | False | v1 | v4 | True |
| v4 | | v2 | v3 | True | v2 | v3 | True |
| | | v2 | v4 | False | v2 | v4 | False |
| | | v3 | v4 | True | v3 | v4 | False |

The relation *Vehicle* in Table 6 contains four vehicle entities (v1 ~ v4). These entities can be used to develop possible combinations of two vehicles interacting with each other as shown data in the first and second column in the relations *Communicate* and *Meet* (i.e., {{v1, v2}, {v1, v3}, {v2, v3}, {v1, v4}, {v2, v4}, {v3, v4}}). The relation *Communicate* means the possible combinations between the two vehicles communicating with each other and contains an attribute *Communicate* indicating whether the two vehicles are communicated (True) or not (False). From data in the relation *Communicate* in Table 5, the true cases for the attribute *Communicate* in the relation *Communicate* in Table 6 can be derived. The true cases for the attribute *Meet* in the relation *Meet* in Table 6 are also derived using the same approach. Now, as we can see Table 6, the relations *Communicate* and *Meet* explicitly contain the true and false cases for the attributes *Communicate* and *Meet*, respectively.

To construct the set of combination between the four vehicles in the relation *Vehicle*, we can use the following script.

**SQL script 4.3**: Combination between the four vehicles
1  **CREATE TABLE**
2      All_Vehicles **AS**
3  ( **SELECT**
4      t1.VID **AS** VID1,
5      t2.VID **AS** VID2
6  **FROM** vehicle **AS** t1
7  **JOIN** vehicle **AS** t2
8  **ON** t1.VID < t2.VID)

The above script generates a new relation called *All_Vehicles*. The dataset for the relation *All_Vehicles* contains {{v1, v2}, {v1, v3}, {v2, v3}, {v1, v4}, {v2, v4}, {v3, v4}}. The above script selects the set of combination between the four vehicles occurring only once. To generate the dataset for the relation *Communicate* in Table 6, we can use the following script.

**SQL script 4.4**: For the Relation *Communicate* in Table 6
1  **SELECT DISTINCT** t2.VID1, t2.VID2,
2  (**SELECT**
3      **IF**(t1. VID1 = t2.VID1 && t1. VID2 = t2.VID2, "True", "False")
4  ) **AS** Communicate
5  **FROM** All_Vehicles t1, Communicate t2

The above script compares data between the relations *All_Vehicles* and *Communicate*. If there is a same primary key between them, a value *True* is assigned to an attribute *Communicate*. If not, a value *False* is assigned to the attribute *Communicate*. To generate the dataset in the relation *Meet* in Table 6, we can use the same approach.

For the relations in Table 6, we assume the following CLD 4.5 in which the meeting between two vehicles may influence the event for communication between the vehicles (i.e., P(*Communicate | Meet*)). In CLD 4.5, $CPC_1$ (Line 1) indicates a condition where two vehicles meet. $CPC_2$ (Line 3) indicates a condition where two vehicles don't meet. For example, $CSD_2$ (Line 4) represents the probability that two vehicles *VID1* and *VID2* communicate with each other in the situation where the two vehicles are not nearby.

**CLD 4.5 [Predicate RV]:** Communicate (*VID1, VID2*)
1  $CPC_1$:   **if** some *VID1.VID2* **have** (Meet = *True*) [
2  $CSD_1$:        *True* = $\Theta_{1.1}$, *False* = $\Theta_{1.2}$
3  $CPC_2$:   **if** some *VID1.VID2* **have** (Meet = *False*) [
4  $CSD_2$:        *True* = $\Theta_{2.1}$, *False* = $\Theta_{2.2}$
5              ]

To learn parameters in CLD 4.5, CSD datasets for $CPC_1$ and $CPC_2$ are required. To generate such datasets, the processes in Section 4.3.1 *Map to Reasoning Model* can be used. For example, a joined dataset between the relations *Communicate* and *Meet* is generated by matching same vehicle entities in both relations. The joined dataset contains four attributes *VID1*, *VID2*, *Communicate*, and *Meet* (e.g., {{v1, v2, True, True}, …, {v3, v4, True, False}). Then, parameter learning as described in Section 4.3.4 *Parameter Learning* is used to construct the parameters in CLD 4.5 (i.e., P(*Communicate | Meet*)).

## 4.4 Test Reasoning Model

In the *Test Reasoning Model* step, a learned reasoning model is evaluated to determine whether to accept it. The accepted reasoning model is output as a final result in this step. This step is decomposed into two sub-steps (Fig. 8): (1) an *Experiment Reasoning Model* step and (2) an *Evaluate Experimental Results* step.

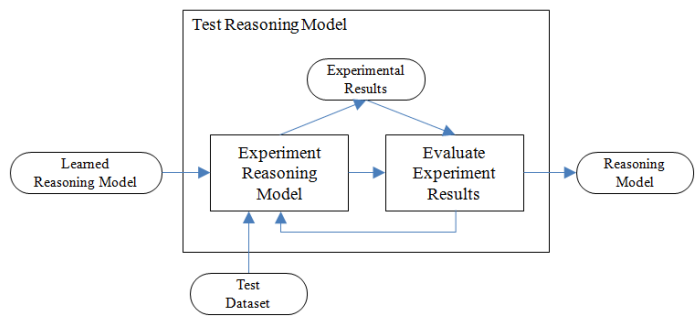

**Fig. 8 Test Reasoning Model**

### 4.4.1 Experiment Reasoning Model

In this section, we introduce

The *Experiment Reasoning Model* step tests the learned reasoning model using a test dataset. The test dataset can be generated from simulations, existing data and/or actual experiments. This experiment can consist of the following five steps. (1) The learned reasoning model is exercised on a test case from the test dataset. (2) The test dataset provides ground truth data to evaluate with a certain metric (e.g., the continuous ranked probability score) in the requirements defined in the *Analyze Requirement* step. (3) The metric is used to measure performance between results from the learned reasoning model and the ground truth data. (4) Steps 1-3 are repeated for all testing cases. (5) This step results in a result value integrating all measured values (e.g., an average of the continuous ranked probability scores).

### 4.4.2 Evaluate Experimental Results

In the *Evaluate Experimental Results* step, the performance of estimation and prediction for the learned reasoning model is assessed by the performance criteria in the requirements defined in the *Analyze Requirement* step (e.g., an average of the continuous ranked probability scores < 0.001). If the measured value satisfies the criteria, the learned reasoning model is accepted and this step results in the learned reasoning model. If the requirement is not satisfied, we can return to the previous steps to improve the performance of the learned reasoning model.

## 4.5 Summary of HML

We introduced a MEBN learning framework, called HML, which contained four steps ((1) Analyze Requirements, (2) Define World Model, (3) Construct Reasoning Model, and (4) Test Reasoning Model). The following list shows their specific sub-steps.

*(1) Analyze Requirements*

    *(1.1) Identify Goals*

    *(1.2) Identify Queries/Evidence*

    *(1.3) Define Performance Criteria*

*(2) Define World Model*

    *(2.1) Define Structure Model*

    *(2.2) Define Rules*

        *(2.2.1) Define Causal Relationships between RVs*

        *(2.2.2) Define Distributions of RVs*

*(3) Construct Reasoning Model*

    *(3.1) Map to Reasoning Model*

        *(3.1.1) Perform Entity-Relationship Normalization*

        *(3.1.2) Perform MEBN-RM Mapping*

        *(3.1.3) Update Reasoning Model using the Rules*

            *(3.1.3.1) Join Relations*

            *(3.1.3.2) Construct Input/Parent Nodes*

            *(3.1.3.3) Construct Context Nodes*

            *(3.1.3.4) Refine Context Nodes*

    *(3.2) Learn Reasoning Model*

*(4) Test Reasoning Model*

    *(4.1) Conduct Experiments for Reasoning Model*

        *(4.1.1) Test Reasoning Model from Test Dataset*

*(4.1.2) Measure Performance for Reasoning Model*

*(4.2) Evaluate Experimental Results*

In (1) the *Analyze Requirements* step, there are three sub-steps: (1.1) the *Identify Goals* step, (1.2) the *Identify Queries/Evidence* step, and (1.3) the *Define Performance Criteria* step. The goals representing missions of the reasoning model is defined in (1.1). The queries, specific questions for which the reasoning model is used to estimate and/or predict answers, and the evidence, inputs used for reasoning, are defined in (1.2). Each query should include performance criteria (1.3) for evaluation of reasoning.

In (2) the *Define World Model* step, there are two sub-steps: (2.1) the *Define Structure Model* step and (2.2) the *Define Rules* step. The *Define Rules* step (2.2) contains two sub-steps: (2.2.1) the *Define Causal Relationships between RVs* step and (2.2.2) the *Define Distributions of RVs* step. In (2.2.1), candidate causal relationships (e.g., *influencing*(*A*, *B*) and *causal*(*A*, *B*)) between RVs are specified using expert knowledge. In (2.2.2), a (conditional) local distribution of an RV is defined by expert knowledge.

In (3) the *Construct Reasoning Model* step, there are two sub-steps: (3.1) the *Map to Reasoning Model* step and (3.2) the *Learn Reasoning Model* step. The *Map to Reasoning Model* step (3.1) is composed of three sub-steps: (3.1.1) the *Perform Entity-Relationship Normalization* step, (3.1.2) the *Perform MEBN-RM Mapping* step, and (3.1.3) the *Update Reasoning Model using the Rules* step. Before applying MEBN-RM to a relational model, the relational model is normalized using Entity-Relationship Normalization (3.1.1). In (3.1.2), MEBN-RM is performed to construct an initial MTheory from the relational model. In (3.1.3), the initial MTheory is updated according to the rules defined in (2.2). The *Update Reasoning Model using the Rules* step (3.1.3) contains four sub-steps: (3.1.3.1) the *Join Relations* step, (3.1.3.2) the *Construct Input/Parent Nodes* step, (3.1.3.3) the *Construct Context Nodes* step, and (3.1.3.4) the R*efine Context Nodes* step. In (3.1.3.1), some relations are joined and an updated MFrag is created, if RVs in a rule are defined in different relations. The causal relationships for the RVs in the rule are defined in the updated MFrag through (3.1.3.2). In (3.1.3.2), if there is an input node, ordinary variables associated with the input node are defined in the updated MFrag. In (3.1.3.3), the context nodes associated with the RVs in the rule are defined in the updated MFrag. For this, the conditions (specified by a "Where" conditioning statement in SQL) in a joining script, used for joining relations in (3.1.3.1), can be reused to construct such context nodes. In (3.1.3.4), ordinary variables sharing the same entity (e.g., IsA (*t*, TIME) and IsA (*t1*, TIME)) are converted into a single ordinary variable (e.g., IsA (*t*, TIME)). Then, equal-context nodes (e.g., $t = t1$) for such ordinary variables are removed. In (3.2) the *Learn Reasoning Model* step, a parameter learning algorithm performs to each RV in the updated MFrag using a training dataset to generate the parameter of the distribution for the RV.

In (4) the *Test Reasoning Model* step, there are two sub-steps: (4.1) the *Conduct Experiments for Reasoning Model* step and (4.2) the *Evaluate Experimental Results* step. In (4.1) there are two sub-steps: (4.1.1) the *Test Reasoning Model from Test Dataset* step and (4.1.2) the *Measure Performance for Reasoning Model* step. In (4.1.1), the learned MTheory from (3) the *Construct Reasoning Model* step is tested using a test dataset and (4.1.2) measured for performance between results from the learned MTheory and the ground truth data in the test dataset. In (4.2) the *Evaluate Experimental Results* step, whether the learned MTheory is accepted or not is decided using the performance criteria defined in (1.3).

In this research, some steps in HML are automated (e.g., (3.1.2) the *Perform MEBN-RM Mapping* step), while some other steps are not yet automated (e.g., (3.1.1) the *Perform Entity-Relationship Normalization* step) but could be automated. Also, some other steps (e.g., (1.1) the *Identify Goals* step) require aid from human (i.e., human centric). The following table shows the level of automation (i.e., *Automated*, *Automatable*, and *Human centric*) for each step in HML.

**Table 7 Processing Method for Steps in HML**

| Main Steps | Sub-steps | Processing Method |
|---|---|---|
| (1) Analyze Requirements | (1.1) Identify Goals | Human centric |
| | (1.2) Identify Queries/Evidence | Human centric |
| (2) Design World Model and Rules | (2.1) Design World Model | Human centric |
| | (2.2) Design Rules | Human centric |
| (3) Construct Reasoning Model | (3.1.1) Perform Entity-Relationship Normalization | Automatable |
| | (3.1.2) Perform MEBN-RM Mapping | Automated |
| | (3.1.3) Update Reasoning Model using the Rules | Automatable |
| | (3.2) Learn Reasoning Model | Automated |
| (4) Test Reasoning Model | (4.1) Conduct Experiments for Reasoning Model | Automatable |
| | (4.2) Evaluate Experimental Results | Automatable |

For example, the (3.2) *Learn Reasoning Model* step is automated by the MEBN-RM mapping algorithm (Section 3.6). The (3.1.1) the *Perform Entity-Relationship Normalization* step is automatable by developing an algorithm converting from ordinary relations to the relations satisfying Entity-Relationship Normalization. The (1.1) *Identify Goals* step is human centric and require human support to perform it. Automatable steps can become automated steps by developing specific processes, algorithms, and software programs. We leave these as future studies.

We developed HML Tool that performs MEBN-RM and the MEBN parameter learning. HML Tool is a JAVA based open-source program that can be used to create an MTheory script from a relational data. This enables rapid development of an MTheory script by just clicking a button in the tool. This is available on Github[5] (see Appendix B).

## 5  Experiment for UMP-ST and HML

We conducted an experiment to compare two MEBN development processes (UMP-ST and HML) in terms of development time supervised by a IRBNet support team (IRBNet ID: 1054232-1). A MEBN model can be constructed by UMP-ST and HML. UMP-ST is the traditional manual approach to develop a MEBN model, while HML is the new approach which is studied in this dissertation.

In this experiment, there were two groups (A and B) selected from six adult people. Both groups were required to develop a MEBN model from stakeholder requirements. The main requirement was to develop a MEBN model for a very simplified domain of a steel plate factory. Thus, we conducted a simplified development experiment, MEBN modelling for simple heating machinery. For the experiment, we tried to constitute same conditions for both groups (e.g., same level of knowledge for a certain domain, BN, MEBN, and MEBN modelling) to draw more general conclusions. Finding and inviting participants who are working for a same domain and have same level of knowledge is difficult. For that reason, the participants who didn't have any experience in the target domain for the experiment were selected and provided domain knowledge to develop a MEBN model. Thus, knowledge for simple heating machinery was given for both groups. This knowledge given to the participants is introduced in Section 5.1.

---

[5] Github is a distributed version control system (https://github.com).

For the experiment, we performed three processes (preparation, execution, and evaluation). In the preparation process, we prepared the experimental settings to make both groups to have same conditions in terms of knowledge and skill for MEBN modelling for the simple heating machinery. In the execution process, the main experiment for MEBN modelling was conducted. In the process, participants in both groups had developed MEBN models using the two methods assigned to each of them. In the evaluation process, development times by the participants were analysed and MEBN models developed by them were tested in terms of accuracy using a simulated test dataset.

## 5.1 Preparation Process

For the experiment, six adult people were invited as subjects. Initially, they didn't have much knowledge about BN and were completely unfamiliar with MEBN, UMP-ST, and HML. (1) Before the execution process, the six people were given such knowledge to develop a MEBN model for the simple heating machinery in the execution process. The lecture contained the minimum amount of knowledge for (continuous) BN, BN modelling, MEBN, a script form of MEBN, and MEBN modelling (i.e., UMP-ST) to develop the MEBN model for the simple heating machinery. Note that a lecture for full knowledge of such domains may require several semesters, so the scope of the experiment was reduced to a smaller size (i.e., the simple heating machinery) rather than the development of a MEBN model for full heating machinery. The six people were divided into two groups (Groups A and B). Group A used UMP-ST, while Group B used HML to develop a MEBN model. (2) To constitute same conditions for each group in terms of skills and domain knowledge for MEBN and UMP-ST, a short test for such knowledge was taken to all of the participants. (3) The short test was graded by a MEBN expert who was not investigator for this experiment. An identity of each participant was not given to the MEBN expert to prevent intervention of prejudice. To penalize our new approach HML, the first (third and fifth) ranked participant belonged to Group A. Second (fourth and sixth) ranked participant belonged to Group B.

**Table 8 Preparation Process**

| Steps | Group A (UMP-ST) | Group B (HML) | Time |
|---|---|---|---|
| 1. Obtain relevant knowledge | Provided a lecture for BN, MEBN, the script form of MEBN, and UMP-ST | | 4 hours |
| 2. Take a short test | Provided a short test for UMP-ST & MEBN | | 30 min |
| 3. Divide into two groups | Graded the test results and selected participants for two groups | | 1 hour |
| 4. Obtain HML knowledge | None | Provided a lecture for HML | Time was checked (*Time A*) |

Before the execution process, (4) HML lecture was provided to Group B. The time for the lecture was checked as *Time A*. The lecture contained the process of HML, the reference PSAW-MEBN model, and how to use the HML tool.

## 5.2 Execution Process

In the execution process, the both groups were requested to develop a MEBN model for a simple heater system heating a slab to support a next manufacturing step (e.g., a pressing step for the slab to make a steel plate). The MEBN model aimed to predict a total cost for the heater given input slabs.

**Table 9 Execution Process**

| Steps | UMP-ST (Group A) | HML (Group B) | Time |
|---|---|---|---|
| 5. Obtain stakeholder requirements and domain knowledge | Provided stakeholder requirements and domain knowledge to both groups | | 1 hour |
| 6. Analyze Requirements | Developed MEBN model Requirements | | Time was checked (*Time B*) |
| 7. Define World Model | Developed a structure model and rules | | Time was checked (*Time C*) |
| 8. Construct Reasoning Model | Developed MEBN model using the script form of MEBN | Develop MEBN model using the HML tool | Time was checked (*Time D*) |

(5) In the first step of the execution process, both groups were given a stakeholder requirement, "Develop a MEBN model which is used to predict a total cost given input slabs". Also, domain knowledge was given to the participants.

The domain knowledge was about the following information: The simple heater system is associated with two infrared thermal imaging sensors to sense the temperature of a slab, each sensor has a sensing error with a normal distribution with a mean zero and a variance three, N(0, 3) (e.g., if it sensed 10 ℃, this means that the actual temperature is in a range between 7.15 ℃ and 12.85 ℃ with the 95th percentile), the heater system contains an actuator which is used to control an energy value to heat a slab (i.e., the actuator calculates the energy value given the input slab temperature), there is no energy loss when the energy value is used in the heater, all manufacturing factors (e.g., the temperature, energy value, and cost) are normally distributed continuous values, the energy unit is kWh (kilowatt-hour), there is a fixed slab weight 100kg, there is an ordered fixed temperature 1200 ℃ for an output slab coming from the heater, and the energy cost is 20cent/kWh.

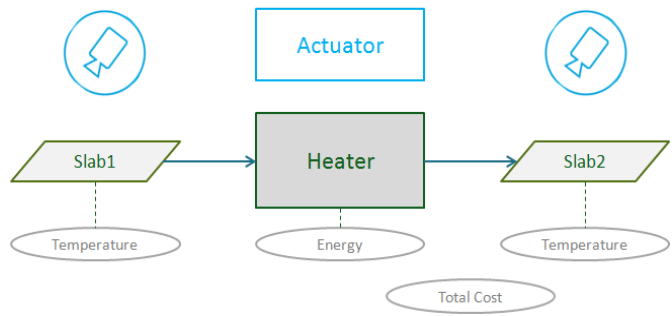

**Fig. 9 Situation for the simple heating machinery**

Also, an idea of how to model the sensor error using BN was given. For example, to include the sensor error, two random variables are used. The first random variable is for an actual temperature, while the second random variable is for a sensed temperature. The actual temperature, then, influences the sensed temperature with the error normal distribution (i.e., N(0, 3)). This can be modelled in a BN as P(*sensed temperature* | *actual temperature*) = *actual temperature* + N(*0, 3*).

For the situation of the simple heating machinery, datasets were generated by a simulator containing a ground truth model designed by a domain expert. The ground truth model contained two parts. The first part is for an actual model which represents a physical world which can't be observed exactly. The second part is for a sensed model which represents an observed world where we can see using sensors. Therefore, the datasets were divided into two parts: Actual data and sensed data (Fig. 10). The sensed data (data sets in the rounded boxes in Fig. 10) were provided to both groups in two formats: The data in an excel format and the data in a relational database (RDB) (Fig. 11). The actual data (e.g., actual temperatures) were not given to either group.

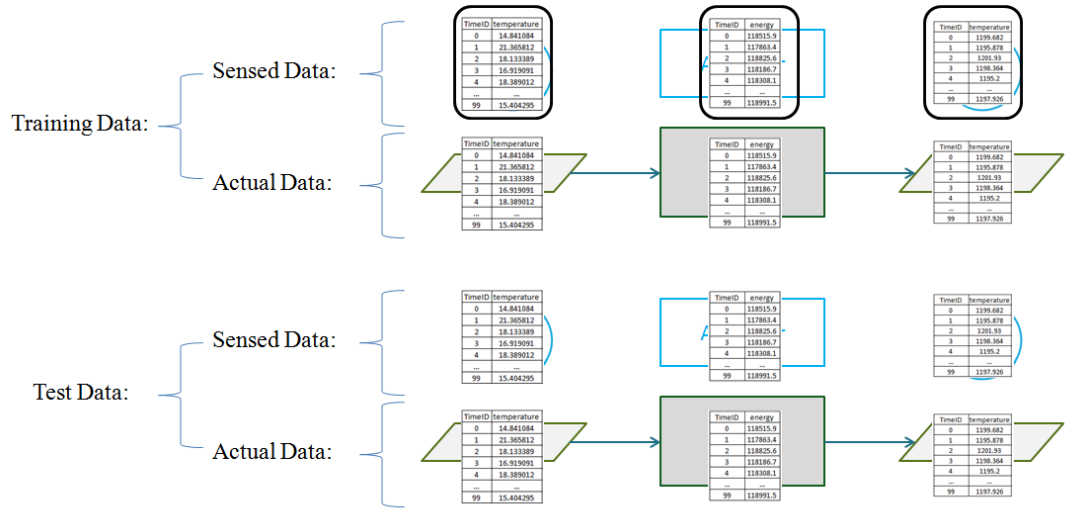

**Fig. 10 Each of training and test data has sensed data and actual data for the simple heating machinery**

Also, the simulator generated two datasets (as shown in Fig. 10): One was a training dataset which was used by the participants to understand the context of the situation and learn a MEBN model using HML, and another was a test dataset which was used to evaluate the models developed by the participants in terms of prediction accuracy for the total cost (6.4.3 Evaluation Process). For this model evaluation, the actual and sensed data in the test dataset were used. For example, sensed data for the temperature of an input slab were used as evidence for the developed model and the developed model was used to reason about a predicted total cost. The predicted total cost was compared with a total cost derived from the actual data in the test dataset.

Participants were requested to (6) develop MEBN model requirements, (7) Define World Model, and (8) construct Reasoning Model. And the development times *Time B*, *Time C*, and *Time D* respectively were checked.

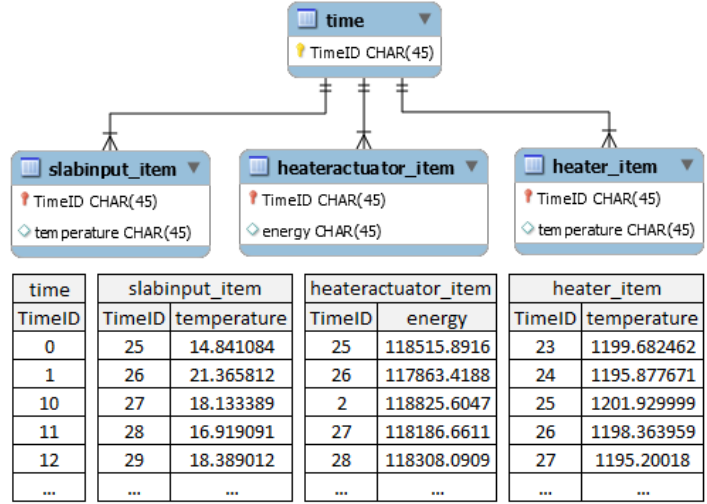

Fig. 11 Sensed datasets for the simple heating machinery

### 5.3 Evaluation Process

In this process, the MEBN models developed by the participants were evaluated and their development times were analyzed. For the model evaluations, simulated test datasets were used. The development times for both were measured according to the use of methods UMP-ST and HML.

Our goal for this experiment is to compare two methods in terms of the development time. However, in some cases, a MEBN model is developed quickly with low accuracy. The comparison between a low quality model and a high quality model in terms of the development time is unfair. Thus, obviously, in order to demonstrate the superiority of HML against UMP-ST, the results of this experiment should ensure two things: A quality of the model developed using HML is equal or better than a quality of the model developed using UMP-ST, and the development time using HML is faster than using UMP-ST. In this experiment, we used the prediction accuracy as the quality of the developed model, because the mission of the developed model is to predict the total cost for the simple heating machinery.

(9) The first step in this process is to evaluate the accuracies of the MEBN models developed by the participants. For this, an accuracy test for the models was performed to determine how well the models predict the total cost using the test dataset generated from the simulator. The simulator generated the test dataset regarding a situation in which three slabs were inputs and a total cost for heating the three slabs was an output. The total cost was calculated using the energy values in the actuator. The output (the total cost) was used to compare a predictive cost reasoned from the MEBN models. To the comparison between the total cost and the predictive cost, we used a continuous ranked probability score (CRPS) in which a perfect prediction yields a score of zero. For prediction accuracy metrics, we can use a mean absolute error (MAE). MAE uses a mean value only to compare an actual (or observed) value with a predictive value, while CRPS uses a predicted probability distribution for comparison (i.e., a mean and a variance). Therefore, CRPS analysis is more precise than MAE analysis. In this step, for a case (i.e., three input slabs and one output total cost) in the test data, CRPS was calculated using a predictive cost. Then, 100 cases were used to compute 100 CRPSs and they were averaged (i.e., Average CRPS).

(10) The development times for both groups were measured according to each step in UMP-ST and HML. The development times *Time A~D* were checked in the preparation process and the execution process. In this step, a total development time was calculated. For Group A, a total development time included *Time B~D*, while for Group B, a total development time included *Time A~D*.

**Table 10 Evaluation Process**

| Steps | UMP-ST (Group A) | HML (Group B) |
|---|---|---|
| 9. Evaluate accuracy of model | Tested both models in terms of accuracy using a simulated test dataset | |
| 10. Evaluate development time | Measured the development times for both according to the use of methods UMP-ST and HML | |

## 5.4 Comparison Results

Table 11 shows an average CRPS to show model accuracy and a total development times to show efficiency of modelling methods for each participant. In the table, the grand CRPS average for Group A is higher than the grand CRPS average for Group B. This means that the MEBN models from Group B are better than the models from Group A in terms of accuracy. Then, the comparison for the total development times makes sense. The average of the total development times from Group A is higher than the average of the total development times from Group B. This implies that HML is a faster process than UMP-ST.

**Table 11 Comparison results**

| Group | Participants | Average CRPS | Total Development Times (Hours: Minutes) |
|---|---|---|---|
| Group A (UMP-ST) | #1 | 1735.3 | 1:06 |
| | #2 | 74.6 | 2:48 |
| | #3 | 114.78 | 2:21 |
| | Grand Average (Standard Deviation) | 641.53 (947.45) | 2:05 (0:52) |
| Group B (HML) | #4 | 45.05 | 1:02 |
| | #5 | 45.05 | 0:58 |
| | #6 | 40.48 | 1:28 |
| | Grand Average (Standard Deviation) | 43.53 (2.64) | 1:09 (0:16) |

In the experiment, we expected the participants would develop an ideal MEBN model. The following figure shows ideal conditional relationships between random variables for the simple heating machinery. In the ideal model, there are three parts: A situation group, an actual target group, and a report group. The situation group contains a random variable representing an overall total cost for this system. The actual target group contains three random variables (a temperature for an input slab, an actual energy for heating, and a temperature for an output slab). The report group contains two random variables (an observed/sensed temperature for the input slab and an observed/sensed temperature for the output slab).

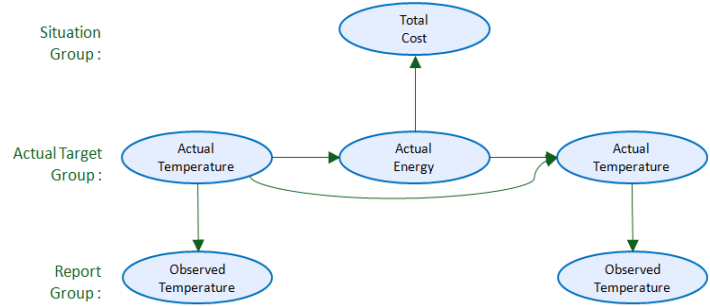

**Fig. 12 Ideal conditional relationships between random variables for the simple heating machinery**

In the experiment, we observed where the participants spent a lot of time. Table 12 shows time-consuming tasks in the experiment. The mark "X" in the table means that it is a time-consuming task for the method.

**Table 12 Comparison results for time-consuming tasks**

| Time-consuming tasks in the experiment | Group A (UMP-ST) | Group B (HML) |
|---|---|---|
| 1. Following process (UMP-ST or HML) | X | - (Supported by HML tool) |
| 2. Finding structure model/rules | X | X (Supported by the PSAW-MEBN reference model) |
| 3. Finding entity/RV/MFrag from relational data | X | - (Supported by MEBN-RM) |
| 4. Finding parameter | X | - (Supported by MEBN parameter learning) |

(1) Following UMP-ST process: Although the participants had studied UMP-ST, it was not easy to follow the process. They didn't have many experiences to develop a MEBN model using UMP-ST, so they were not familiar with the process. They remembered the process by reading a UMP-ST paper and developed their model according to each step of UMP-ST. For Group B, the HML tool supported the development of a MEBN model. By clicking some buttons in the HML tool, each step in HML was shown and the participants could make the models quickly. (2) Finding Structure Model/Rules: The participants in both Groups were required to find the structure model for the simple heating machinery. Although knowledge of the simple heating machinery situation was given, the participants in both groups struggled to find the structure model and rules. Group B was taught about the PSAW-MEBN reference model [Park et al., 2014]. The PSAW-MEBN reference model provides knowledge about a set of random variable groups (Situation, Actual Target, and Report) and causal relationships (i.e., rules) for PSAW. However, such knowledge did not have much influence on the development time for the structure model and rules, because the context for the simple heating machinery was too simple to use the PSAW-MEBN reference model. So, the participants in the two groups thought about their models in similar ways. However, the participants could not be sure whether or not their models were correct, so they spent relatively more time to think about their structure models and rules. (3) Finding entity/RV/MFrag from the RDB: The participants in Group A could not be sure of which elements in the RDB can be entity/RV/MFrag in MEBN, so they used times to figure out this. On the other hand, the participants in Group B used the HML tool containing MEBN-RM, so they didn't consider this step much. (4) Finding CLD: The participants in Group A looked at data to find normal distributions and regression models for RVs, while the participants in Group B used the MEBN parameter learning built in the HML tool.

## 6  Conclusion

In this research, we introduced a new development framework for MEBN, providing a semantically rich representation that also captures uncertainty. MEBN was used to develop Artificial Intelligence (AI) systems. MEBN models for such systems were constructed manually with the help of domain experts. This manual MEBN modeling was labor-intensive and insufficiently agile. To address this problem, we introduced a development framework (HML) combining machine learning with subject matter expertise to construct MEBN models. We also presented a MEBN parameter learning for MEBN. In this research, we conducted an experiment between HML and an existing MEBN modeling process in terms of the development efficiency. In conclusion, HML could be used to develop more quickly a MEBN model than the existing approach. Future steps for HML are to apply it to realistic Artificial Intelligence systems. Also, HML should be more thoroughly investigated in terms of efficiency (agility for the development of a reasoning model) and effectiveness (producing a correct reasoning model).

## Appendix A: Bayesian Network Learning

Bayesian Networks (BN) learning from data is a process to find a Bayesian network that fits data well. Given a graph of BN and a dataset, *Parameter Learning* is the problem of finding a parameter $\theta$ that provides a good fit to the data. *Structure Learning* is the problem of finding a graph $G$ of BN that provides a good fit to the data. *Structure Learning* can have following topics: (1) dependency or independency between nodes can be learned, (2) a hidden or an unobserved node in a BN can be found, and (3) a functional form of a local distribution for a node can be identified.

In Bayesian theory, any uncertain aspect of the world can be represented as a random variable (RV). In BN learning, data $D$, graph $G$, and parameter $\theta$ can be RVs. The set of data $D$, graph $G$, and parameter $\theta$ are represented as $\boldsymbol{D}$, $\boldsymbol{G}$, and $\Theta$, respectively. For BN, data $D$ means flat data which has no relationships among its records, while for MEBN learning in this research, data $D$ means relational data. In the following subsections, we introduce common approaches for BN parameter learning. We refer to [Pearl 1988][Heckerman, 1998][Koller & Friedman, 2009] for following subsections.

### A.1.  BN Parameter Learning

BN parameter learning is to find parameters that fit a dataset well. We can think of this as inference in probability theory. Two of the most popular approaches to estimating parameters from data are *Maximum Likelihood Estimation* (MLE) and *Bayesian inference*. In the Bayesian approach, we begin with a prior distribution for the parameter and use the data to obtain a posterior distribution. MLE finds the parameter that maximizes the likelihood function, and does not use a prior distribution. The use of the prior distribution in the Bayesian approach can help to overcome the over-fitting problem of learned parameters. We introduce MLE first and then the Bayesian approach.

### A.1.1.  Maximum Likelihood Estimation

For MLE, let's assume that we are doing a statistical experiment (e.g., tossing coins, where H is a head and T is a tail). In the experiment, there is a set of independent and identically distributed (IID) observations $\boldsymbol{D} = \{D_1, D_2, \ldots, D_n\}$ (e.g., {H, H, T}), which are drawn at random from a distribution with an unknown probability density or mass function $f(D_i \mid \theta_A)$, where $n$ is the number of the observations and $\theta_A$ is an actual parameter for the distribution. Since we don't know the actual parameter $\theta_A$, we find an estimator $\theta^*$ which would be close to the actual parameter $\theta_A$. For this, we introduce a function, called *likelihood*, which is a function of a parameter $\theta$ for given observations $D$ and is used to find the estimator $\theta^*$.

$$L(\theta : D) = P(D \mid \theta), \tag{A.1}$$

where $D$ is the observations and $\theta$ is a parameter.

The parameter $\theta$ can contain sub-parameters $\{\theta_1, \theta_2, \ldots, \theta_m\}$, where $m$ is the number of the sub-parameters in $\theta$. For example, the parameter $\theta$ for a normal distribution can contain two sub-parameters $\theta_1$ (for mean) and $\theta_2$ (for variance). Also, a parameter $\theta$ can have only one sub-parameter $\theta_1$ (e.g., $\theta_1 = P(H)$).

The likelihood function $L(\theta : D)$ in Equation A.1 is equal to the probability of the observations $D$ given a parameter $\theta$. We then define an estimator $\theta^*$, called a *maximum likelihood estimator* in Equation A.2.

$$\theta^* = arg\ max_{\theta \in \Theta} L(\theta : D). \tag{A.2}$$

where $\Theta$ is a set of parameters.

Equation A.2 means that in the set of parameters, a maximum likelihood estimate or parameter which maximizes the likelihood function is found. We can think of various types of distribution for the observations $D$ in the experiment. If we consider an RV X for the observations $D$ with the multinomial distribution, we can have the following equation which is the maximum likelihood estimator for the parameter $\theta$.

$$\theta_k^* = \frac{C[x_k]}{\sum_{q=1}^{N} C[x_q]}, \tag{A.3}$$

where C[.] is a function returning the number of times a value $x_k \in Val(X)$ in an RV X appears in $D$ and $N = |Val(X)|$. Note that for a variable X, a function Val(X) returns a set of values for X.

For example, suppose that there is an RV X for an observation $D_i$. The RV X contain two values $x_1 = H$ and $x_2 = T$ and there is a set of observations $D = \{H, H, H, T\}$. On a set of observations, the count of the number for $x_1$ and $x_2$ is observed using the function C[.]. For example, $C[x_1] = 3$ and $C[x_2] = 1$. Using Equation A.3, we can calculate the maximum likelihood estimates for $x_1$ and $x_2$ as $\theta_1^* = 3/4$ and $\theta_2^* = 1/4$, respectively.

We can use MLE for a Bayesian network (BN) to estimate a parameter of an RV in the BN. Suppose that there is an RV $X_i$ in the BN, $x_k$ is a value for the RV $X_i$ (i.e., $x_k \in Val(X_i)$), there is a set of parent RVs for the RV (i.e., $Pa(X_i) = U$), and u is some instantiation for the set of parent RVs (i.e., $u \in Val(U)$). If we assume that each RV $X_i$ is the multinomial distribution and the observations associated with the RV $X_i$ are independent and identically distributed, then the maximum likelihood estimator for a value $x_k|u$ in the RV $X_i$ in the BN can be formed as Equation A.4.

$$\theta^*_{i\,x_k|u} = \frac{C[x_k, u]}{\sum_{q=1}^{n} C[x_q, u]},\qquad(A.4)$$

where $C[x_q, u]$ is the number of times observation $x_q$ in X and its parent observation u in Val(U) appears in $\boldsymbol{D}$.

For example, we assume that there are a node $X_1$ in a BN, Val($X_1$) = {$x_1$ = T, $x_2$ = F}, the set of parent nodes for $X_1$ (i.e., Pa($X_1$) = U = {$U_1$}), and Val($U_1$) = {$u_1$ = A, $u_2$ = B}. Also, there is a desta set $\boldsymbol{D}$ = {$D_1$ = {T, A}, $D_2$ = {T, A}, $D_3$ = {F, A}}, where the first value in $D_k$ is for $X_1$ and the second value in $D_k$ is for $U_1$. If u = $u_1$ = A, then $\theta^*_{1\,x_1|u_1}$ = 2/3.

### A.1.2. Bayesian Parameter Estimation

For Bayesian approach, we assume that the parameter $\theta$ in the statistical experiment in Section A.1 is a value of an RV $\boldsymbol{\Theta}$ (i.e., P($\boldsymbol{\Theta}$ = $\theta$) = g($\theta$)). In this setting, we try to draw inference about the RV $\boldsymbol{\Theta}$ given a set of IID observations $\boldsymbol{D}$ = {$D_1$, $D_2$, …, $D_n$}, where $n$ is the number of the observations. We then find a *posterior distribution* of the parameter $\theta$ given the observations $\boldsymbol{D}$ (i.e., P($\theta$ | $\boldsymbol{D}$)). For this, we can use Bayes' theorem, shown in Equation A.5. Note that the posterior distribution can be used to compute a *posterior predictive distribution* (or simply a predictive distribution) which is the distribution of a future observation given past observations. The posterior predictive distribution will be discussed later. The following equation shows the posterior distribution.

$$P(\theta \mid \boldsymbol{D}) = \frac{P(\boldsymbol{D} \mid \theta)P(\theta)}{P(\boldsymbol{D})},\qquad(A.5)$$

where $\boldsymbol{D}$ is the current observations and P($\boldsymbol{D}$) > 0.

Bayesian inference computes the posterior distribution P($\theta$ | $\boldsymbol{D}$) using a prior probability P($\theta$) and a likelihood function P($\boldsymbol{D}$ | $\theta$) in Bayes' theorem. The prior probability P($\theta$) is the probability of a parameter $\theta$ before the current observations $\boldsymbol{D}$ are observed. The likelihood function P($\boldsymbol{D}$ | $\theta$) (Equation A.1) is the probability of the observations $\boldsymbol{D}$ given the parameter $\theta$. In Equation A.5, P($\boldsymbol{D}$) is a marginal likelihood (or a normalizing constant) which is the probability distribution for the observations $\boldsymbol{D}$ integrated over all parameters (i.e., P($\boldsymbol{D}$) = $\int_\theta$ P($\boldsymbol{D}$ | $\theta$)P($\theta$)d$\theta$). The posterior distribution P($\theta$ | $\boldsymbol{D}$) is updated using the prior probability P($\theta$) and the likelihood function P($\boldsymbol{D}$ | $\theta$), and this update can be repeated. For example, after applying some observations to Equation A.5, we can have a posterior distribution. This posterior distribution can be regarded as a prior probability. And then given some new observations, we can compute a new posterior distribution.

For the parameter $\theta$, we can have a hyperparameter α which influences the parameter $\theta$ and can be formed as $\theta$ ~ P($\theta$ | α), where the hyperparameter α can be a vector or have sub-hyperparameters. In this setting, the probability for the parameter P($\theta$) can be changed to the probability given the hyperparameter P($\theta$ | α). By adding the hyperparameter to Equation A.5, Equation A.6 is derived under some assumptions that (1) the observations are independent of a hyperparameter given the parameter associated with the hyperparameter and (2) the sample space for parameters is a partition.

$$P(\theta \mid \boldsymbol{D}, \alpha) = \frac{P(\boldsymbol{D} \mid \theta)P(\theta \mid \alpha)}{\int_\theta P(\boldsymbol{D} \mid \theta)P(\theta \mid \alpha) \, d\theta}. \tag{A.6}$$

We use this posterior distribution containing the hyperparameter (Equation A.6) to compute the predictive distribution. The predictive distribution is the distribution of a new observation given past observations.

$$P(D_{new} \mid \boldsymbol{D}, \alpha) = \int_\theta P(D_{new} \mid \theta)P(\theta \mid \boldsymbol{D}, \alpha) d\theta, \tag{A.7}$$

where $D_{new}$ is a new observation and is independent of the past IID observations $\boldsymbol{D}$ given a parameter $\theta$.

In Equation A.7, the predictive distribution integrates over all parameters for the new observation and the posterior distributions (Equation A.6). To compute the predictive distribution (Equation A.7), we should deal with the posterior distribution (Equation A.6) first. If there is no closed form expression for the integral in the denominator in Equation A.6, we may need to approximate the posterior distribution. If there is a closed form expression for the integral in the denominator and, the prior distribution and the likelihood are a conjugate pair, then an exact posterior distribution can be found.

A probability distribution in the exponential family (e.g., normal, exponential, and gamma) has a conjugate prior [Gelman et al, 2014]. We can consider an RV X with a categorical probability distribution. For such a categorical probability distribution, Dirichlet conjugate distribution is commonly used. Using Dirichlet distribution, the predictive distribution will be a compact form [Koller & Friedman, 2009].

$$P(D_{new} \mid \boldsymbol{D}, \alpha) = \frac{\alpha_k + C[x_k]}{\sum_j \alpha_j + \sum_{q=1}^{N} C[x_q]}, \tag{A.8}$$

where C[.] is a function returning the number of times a value $x_k \in$ Val(X) in a variable X appears in $\boldsymbol{D}$ and N = |Val(X)|, α is a hyperparameter, and $\alpha_j$ is a sub-hyperparameter in Dirichlet distribution as shown the following.

$$\theta \sim \text{Dirichlet}(\alpha_1, \alpha_2, \dots, \alpha_N) \text{ if } P(\theta) \propto \prod_{j}^{N} \theta_j^{\alpha_j - 1}, \tag{A.9}$$

where the sub-hyperparameter $\alpha_j$ is the number of samples which have already happened [Koller & Friedman, 2009].

The Bayesian approach above can be used for BN parameter learning. If a prior distribution for an RV $X_i$, $P(\theta^i \mid \alpha)$, is the Dirichlet prior with a hyperparameter $\alpha = \{\alpha_{x_1|u}, \dots, \alpha_{x_N|u}\}$, then the Dirichlet posterior for $P(\theta^i | \alpha)$ is $P(\theta^i | D, \alpha)$ with a hyperparameter $\alpha = \{\alpha_{x_1|u} + C[x_1, u], \ldots, \alpha_{x_N|u} + C[x_N, u]\}$, where a value $x_k \in Val(X_i)$, $u \in Val(Pa(X_i) = U)$, and $C[x_q, u]$ is the number of times observation $x_q$ in $X_i$ and its parent observation u in Val(U) appears in *D*. Using the Dirichlet posterior, we can derive the predictive distribution for a value of $X_i$ in a BN under some assumptions: (1) local parameter independences and (2) global parameter independences [Heckerman et al., 1995].

$$P(X_i = x_k | U = u, D, \alpha) = \frac{\alpha_{x_k|u} + C[x_k, u]}{\sum_{q=1}^{N}(\alpha_{x_q|u} + C[x_q, u])}, \quad (A.10)$$

where $N = |Val(X_i)|$.

Equation A.10 shows the posterior predictive distribution for the value $x_k$ of the *i*-th RV $X_i$ in the BN given a parent value u, the observations *D*, and a hyperparameter α for Dirichlet conjugate distribution.

## Appendix B: Bayesian Network Learning

We developed HML Tool (Fig. B.1) that performs MEBN-RM and the MEBN parameter learning. HML Tool is a JAVA based open-source program[6] that can be used to create an MTheory script from a relational schema. This enables rapid development of an MTheory script by just clicking a button in the tool. The current version of HML Tool uses MySQL. The most recent version and source codes of HML Tool are available online at https://github.com/pcyoung75/GMU_HMLP.git. HML Tool codes are in the GMU_HMLP Github repository.[7]

MEBN-RM Tool which contains three panels: (1) a left tree panel shows a list of relational database, (2) a right top panel shows a result MTheory script, and (3) a right bottom panel shows an input window in which we can insert some information. The following figure shows the interface of HML Tool and a result MTheory script using the tool.

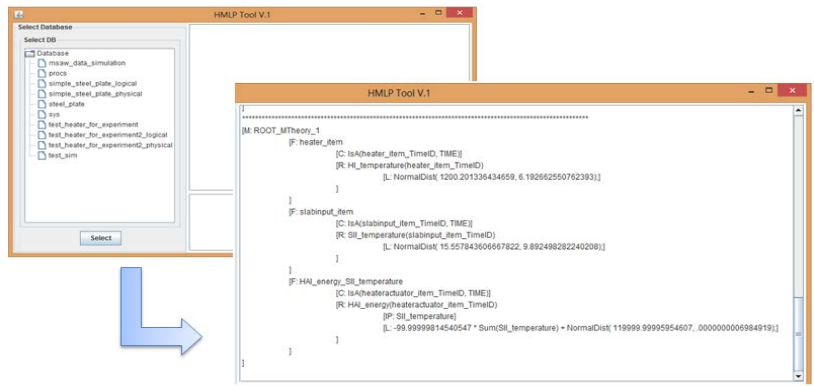

**Fig. B.1 HML Tool**

---

[6]Researchers around the world can debug and extend MEBN-RM Tool.

[7]Github is a distributed version control system (https://github.com).

**Acknowledgements**

The research was partially supported by the Office of Naval Research (ONR), under Contract#: N00173-09-C-4008. We appreciate Dr. Paulo Costa and Mr. Shou Matsumoto for their helpful comments on this research.